\documentclass[10pt,twocolumn,letterpaper]{article}
\usepackage[pagenumbers]{cvpr}
\usepackage{graphicx}
\usepackage{scalerel}
\usepackage{amsmath}
\usepackage{amssymb}
\usepackage{booktabs}
\usepackage[title]{appendix}
\usepackage{tablefootnote}
\usepackage{amsmath,array,xcolor}
\usepackage[referable]{threeparttablex}
\usepackage{amsmath,mathtools}
\usepackage{amssymb}
\usepackage{amsthm}
\usepackage{amsfonts}
\usepackage{times}

\usepackage{footnote}
\makesavenoteenv{tabular}
\makesavenoteenv{table}

\usepackage{epsfig}
\usepackage{graphicx}
\usepackage{amsmath}
\usepackage{amssymb}
\usepackage{float}
\usepackage{caption}
\usepackage{textcomp}
\usepackage{multirow}
\usepackage{amssymb}
\usepackage{makecell}
\usepackage{notoccite}
\usepackage{comment}
\usepackage{subfiles}
\usepackage{threeparttable}
\definecolor{OliveGreen}{rgb}{0,0.6,0}
\definecolor{BgWhite}{rgb}{1,1,1} 
\definecolor{Gray}{rgb}{0.4,0.4,0.4} 
\definecolor{mygray}{gray}{0.6}

\newcommand{\gray}[1]{{\color{Gray}#1}}

\usepackage[pagebackref,breaklinks,colorlinks]{hyperref}
\usepackage[capitalize]{cleveref}
\crefname{section}{Sec.}{Secs.}
\Crefname{section}{Section}{Sections}
\Crefname{table}{Table}{Tables}
\crefname{table}{Tab.}{Tabs.}

\begin{document}

\title{
Lepard: Learning partial point cloud matching in rigid and deformable scenes }

\author{
Yang Li${}^\text{1} $\quad\quad Tatsuya Harada${}^\text{1,2}$ \vspace{0.2cm} \\
${}^\text{1}$The University of Tokyo, \quad ${}^\text{2}$RIKEN\\
{\tt\small \{liyang,harada\}@mi.t.u-tokyo.ac.jp}
}

\newcommand\blfootnote[1]{%
  \begingroup
  \renewcommand\thefootnote{}\footnote{#1}%
  \addtocounter{footnote}{-1}%
  \endgroup
}

\twocolumn[{%
	\renewcommand\twocolumn[1][]{#1}%
	\maketitle
	\begin{center}
        \centering
		\captionsetup{type=figure}
		\includegraphics[width=1\linewidth]{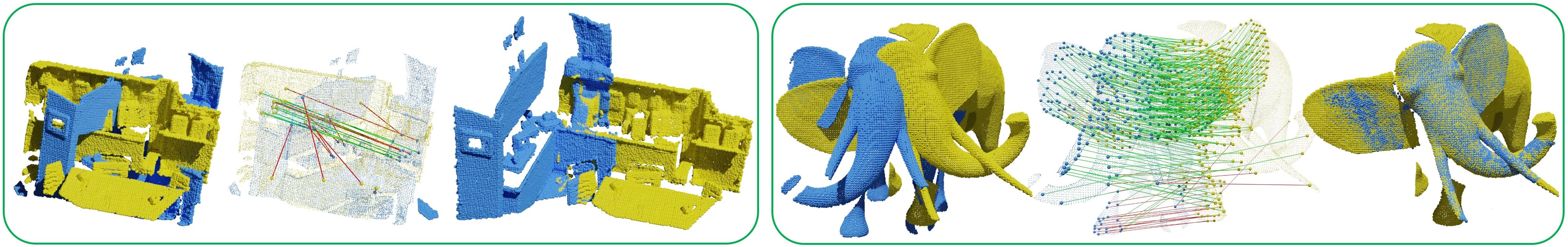}
		\captionof{figure}{
We propose a feature matching method for rigid (left) and deformable (right) point clouds that are captured by range sensors. 
Results shown are the initial alignment, the predicted matches (blue/red lines indicate inliers/outliers), and the registration results. 
Registration is done by RANSAC for rigid and non-rigid ICP for deformable.
}
		\label{fig:teaser}
	\end{center}  
}]

\begin{abstract}
We present Lepard, a \textbf{Le}arning based approach for \textbf{pa}rtial point cloud matching in \textbf{r}igid and \textbf{d}eformable scenes.
The key characteristics are the following techniques that exploit 3D positional knowledge for point cloud matching:  
1) An architecture that disentangles point cloud representation into feature space and 3D position space.
2) A position encoding method that explicitly reveals 3D relative distance information through the dot product of vectors.
3) A repositioning technique that modifies the cross-point-cloud relative positions.
Ablation studies demonstrate the effectiveness of the above techniques. 
In rigid cases, Lepard combined with RANSAC and ICP demonstrates state-of-the-art registration recall of \textbf{93.9\%} / \textbf{71.3\%}  on the 3DMatch / 3DLoMatch.
In deformable cases, Lepard achieves \textbf{+27.1\%} / \textbf{+34.8\%} higher non-rigid feature matching recall than the prior art on our newly constructed 4DMatch / 4DLoMatch benchmark. 
Code and data are available at \href{https://github.com/rabbityl/lepard}{https://github.com/rabbityl/lepard}.
\end{abstract}
\section{Introduction}

Matching partial point clouds from range sensors lies at the core of many 3D computer vision applications including SLAM and dynamic tracking and reconstruction. The former assumes rigid scenes, e.g.~\cite{kinectfusion-Izadi,kinectfusion-Newcombe}, while the latter focuses on scenes that are non-rigidly deforming, e.g.~\cite{dynamicfusion}. 
This work aims at developing a robust point clouds matching method for both rigid and deformable scenes.

Point cloud matching methods often consist of two phases:  point cloud feature extraction followed by nearest neighbor search in feature space.
Recent learning-based works have made substantial progress for representation learning in 3D data.
State-of-the-art point clouds matching approaches~\cite{huang2021predator,d3feat,FCGF} employ the geometry features extracted by 3D convolutional networks, such as the KPConv-based~\cite{thomas2019kpconv} or the Minkowski Engine~\cite{choy2019minkowski}.
These 3D feature extractors are strictly translation invariant, and, to a certain extent, also invariant to rotation transformations given the commonly adopted max-pooling layers in the networks and random rotation-based data augmentation during training.

Transformation invariance is well suited for local geometry feature representation.
However, it may cause ambiguity in scenes that have repetitive geometry patterns.
For instance, the same kind of chairs scattered in different locations of a floor, or left and right hands of a human could yield similar geometry features.
We argue that such ambiguity can be resolved by enhancing geometry features with the 3D positional knowledge. 
Intuitively, humans associate things across observations by referring to not only things' appearance but also their relative locations.

Motivated by the above observations, we design Lepard,  a novel partial point clouds matching method that exploits 3D positional knowledge.
We first build our baseline using the fully convolutional feature extractor KPFCN~\cite{thomas2019kpconv}, the concept of Transformer~\cite{attention_is_all_u_need} with self and cross attention, and the idea of differentiable matching~\cite{sarlin2020superglue, sun2021loftr}.
Then, to leverage 3D position information, we introduce the following techniques:
1) A framework that fully disentangles the point cloud representations into a features space and a position space.
2) A position encoding method that explicitly reveals 3D relative distance information through the dot product of vectors.
3) A repositioning module that adjusts the cross-point-cloud relative positions which benefits cross attention and differentiable matching.
Ablation studies demonstrate the effectiveness of the above techniques.

In addition, we propose a partial point cloud matching benchmark called 4DMatch, and its low overlap version 4DLoMatch.
4DMatch contains point clouds that are non-rigidly deforming across the time axis.  
Compared to the rigid situations, the time-varying geometry in 4DMatch poses more challenges for both matching and registration.

We apply Lepard for both rigid and deformable point cloud matching.
In rigid cases, Lepard combined with RANSAC and ICP demonstrates state-of-the-art registration recall of \textbf{93.9\%} / \textbf{71.3\%}  on the 3DMatch / 3DLoMatch. 
On the newly proposed 4DMatch and 4DLoMatch benchmarks, Lepard achieves \textbf{+27.1\%} and \textbf{+34.8\%} higher non-rigid matching recall than the prior art.

\section{Related work}

\subsection{Rigid Point Cloud Matching and Registration}
Local descriptors prediction followed by the robust RANSAC-based~\cite{schnabel2007efficientRANSAC, holz2015registration} optimization is a long-studied approach.
Early works employ hand engineered descriptors~\cite{fpfh,tombari2010unique,spin_image_3d,rusu2008aligning}. 
Recent learning-based approaches have made significant progresses for point feature representation~\cite{FCGF,huang2021predator,d3feat,3dmatch,3DSN,ppfnet,Ppf-foldnet,ao2021spinnet,YOHO,BYOC,khoury2017LCGF,yang2021sgp,liu2021neighborhood, elbanani2021unsupervisedrr,Multi_view_des2020,yu2021cofinet}. 
3DMatch~\cite{3dmatch} made the first attempt to extract descriptors using the siamese network. 
FCGF~\cite{FCGF} leverages the fully convolutional network~\cite{fcn} structure for dense feature extraction. 
D3feat~\cite{d3feat} jointly learns feature description with point saliency detection.
Predator~\cite{huang2021predator} adopts the attention mechanism to predict overlapping regions for feature sampling.
CoFiNet~\cite{yu2021cofinet} learns feature descriptors in a coarse-to-fine manner.

Another line of research focuses on direct registration.
ICP~\cite{ICP-besl1992} and FGR~\cite{zhou2016FGR} optimize the pose using second-order gradients. Go-ICP~\cite{yang2015goicp} achieves global registration with a SE(3) space searching schema.
Recent works incorporate learned models into end-to-end pose optimization~\cite{3DRegnet, yew2020rpmnet, choy2020DGR, aoki2019pointnetlk,OMNet,Positional_Equilibrium,li2021pointnetlk2}.
PointNetLK~\cite{aoki2019pointnetlk,li2021pointnetlk2} formulate point cloud registration as a Lucas Kanade-based~\cite{inverse_composition_method} optimization task;
Wang et al.~\cite{wang2019dcp,wang2019prnet} learn registration through graph neural networks.
DGR~\cite{choy2020DGR} and 3DRegNet~\cite{3DRegnet} learn correspondence weighting networks to reject outliers.

This paper is about enhancing the point cloud feature descriptors with 3D positional knowledge.

\subsection{Non-Rigid Correspondence} 
Estimating non-rigid correspondence from real-world sensor data is a key task for online non-rigid reconstruction~\cite{dynamicfusion,volumedeform,deepdeform,surfelwarp,splitfusion}.
DynamicFusion~\cite{dynamicfusion} employs the simple projective correspondence for real-time efficiency.
VolumeDeform~\cite{volumedeform} incorperates SIFT~\cite{SIFT} descriptor-based correspondence for robust non-rigid tracking.
Schmidt et al.~\cite{schmidt2016self} uses DynamicFusion to supervise a siamese network for dense correspondence learning.
DeepDeform~\cite{deepdeform} learns sparse global correspondence for patches in non-rigid deforming RGB-D sequences.
Li et al.~\cite{Learning2Optimize} learns non-rigid features through a differentiable non-rigid alignment optimization.
NNRT~\cite{bovzivc2020neural} focuses on end-to-end robust correspondence estimation with an outlier rejection network.
Scene flow estimation, e.g.~\cite{flownet3d,wu2019pointpwc,puy2020flot,Neural_Scene_Flow_Prior}, is a closely related technique which usually delivers inter-frame level correspondence.

Non-Rigid correspondence is also a major topic in geometry processing where the input data are usually manifold surfaces.  
Huang et al.~\cite{isometric_deformations} filters outliers using isometric deformation assumption.
3DCODED~\cite{3d-coded} achieve shape correspondence through latent code optimization.
Functional map~\cite{ovsjanikov2012functionalmaps} have been proposed to produce shape correspondence in~\cite{rodola2017partial,melzi2019zoomout,deepgeofm2020,DPFM}. 

This paper is about developing a general non-rigid feature matching method for partial point cloud scans.

\begin{figure*}
    \centering
    \includegraphics[width=.9\linewidth]{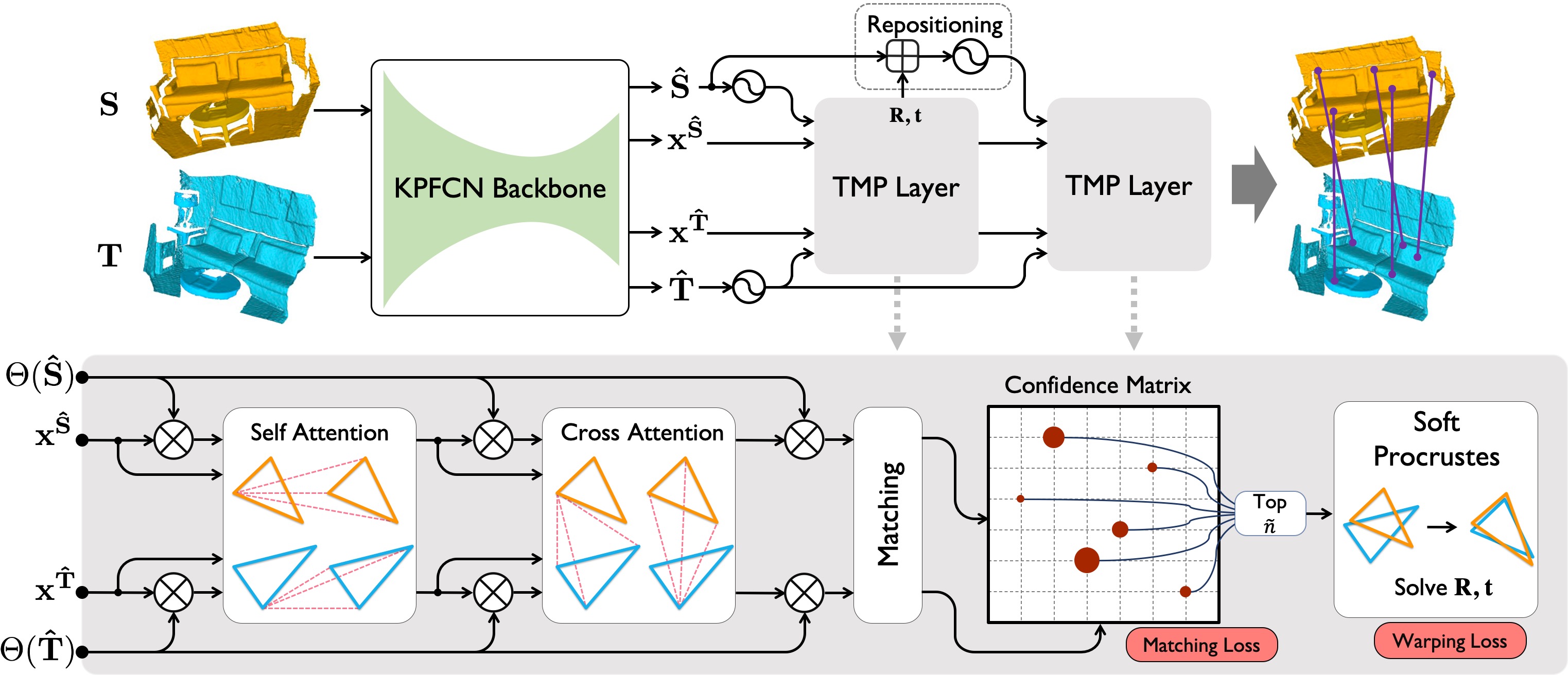}
    \caption{
\textbf{Overview of the proposed method.}
(The symbols are 
\includegraphics[height=0.7em]{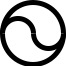} 
: Positional encoding function $\Theta(\cdot)$;
\includegraphics[height=0.7em]{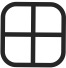}: 
Rigid 3D transformation;
\includegraphics[height=0.7em]{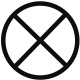}: 
Matrix-vector multiplication;
\includegraphics[height=0.7em]{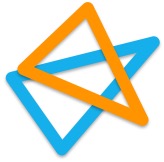}: 
Source and target).
Given the input point cloud $\mathbf{S}$ and $\mathbf{T}$, the KPFCN backbone grid-subsample them to $\mathbf{\hat{S}}$ and $\hat{\mathbf{T}}$, and extract geometry features $\mathbf{x}^\mathbf{\hat{S}}$ and $\mathbf{x}^\mathbf{\hat{T}}$ (Sec.~\ref{sec:backbone}). 
The positions informations are encoded as $\Theta (\mathbf{\hat{S}})$ and $\Theta (\mathbf{\hat{T}})$ using the 3D relative positional encoding function (Sec.~\ref{sec:3d_pe}).
The position codes and geometry features are then processed by the first \textbf{TMP} layer which includes a \textbf{T}ransformer block with self and cross attentions (Sec.~\ref{sec:transformer}), a differentiable \textbf{M}atching layer (Sec.~\ref{sec:matching}), and a soft \textbf{P}rocurestes layer to estimate the rigid fitting $\mathbf{R,t}$ (Sec.~\ref{sec:prucrustes}).  Based on the rigid fitting estimation, the Repositioning layer adjusts source's position code $\Theta (\mathbf{\hat{S}})$ (Sec.~\ref{sec:Repositioning}). Given the updated positions and transformed features, the second TMP layer predicts the final matches. }
    \label{fig:pipeline}
\end{figure*}

\section{Problem Definition}
Given a source point clouds $\mathbf{S} \in \mathbb{R}^{n\times3}$ and a target point cloud $\mathbf{T} \in \mathbb{R}^{m\times3}$, where $n,m$ are the number of points, 
our goal is to find a set of matches $\mathcal{K}$, which can be used to recover the warp function $\mathcal{W}: \mathbb{R}^3 \mapsto \mathbb{R}^3$ that aligns $\mathbf{S}$ to $\mathbf{T}$. 
In this paper, we focus on both rigid and deformable point clouds.
In rigid cases, the warp function $\mathcal{W}$ is parameterized by a $ SE(3)$ transformation.
In deformable cases, $\mathcal{W}$ is generalized to the dense per-point warp field.
Given the ground truth warp function $\mathcal{W}_{gt}$,
an inlier match $(\mathbf{S}_i\in \mathbb{R}^3, \mathbf{T}_j \in \mathbb{R}^3)\in \mathcal{K}$ should satisfies 
$
||\mathcal{W}_{gt}(\mathbf{S}_i) - \mathbf{T}_j ||_2  < \sigma
$,
where $||\cdot||_2$ is the Euclidean norm, and $\sigma$ is the tolerance radius for a match. 

\smallskip
\noindent
\textbf{Partial Overlap.}
In real-world range sensor data, due to object motion or viewpoint change, a point in $\mathbf{S}$ does not necessarily have a corresponding point in $\mathbf{T}$. This is referred to as a non-overlapping point. 
We define the set of overlapping points $\mathcal{O}^{\mathbf{T}}_{\mathbf{S}}$ for the source point cloud by:
$$
\mathcal{O}^{\mathbf{T}}_{\mathbf{S}}=\{ \mathbf{S}_i| \mathbf{S}_i \in \mathbf{S}\; \wedge \; ||\mathcal{W}_{gt}(\mathbf{S}_i) - \text{NN}(\mathcal{W}(\mathbf{S}_i), \mathbf{T}) ||_2  < \sigma \}
$$
where $\text{NN}(\cdot,\cdot)$ is the nearest neighbor search operator.
Then the overlap ratio can be calculated by $|\mathcal{O}^{\mathbf{T}}_{\mathbf{S}}|/|\mathbf{S}|$. Fig.~\ref{fig:4dmatch_eg} shows examples of non-rigid point clouds pairs with different overlap ratio.

\section{Method}

Fig.~\ref{fig:pipeline} shows the overview of the proposed method.

\subsection{Local geometry feature extraction}\label{sec:backbone}
Given the input source and target point clouds $\mathbf{S}$ and $\mathbf{T}$,
We use the function $\Phi$ to extract multi-level geometry features. $\Phi$ does the following two mappings:
$$   
(\mathbf{\hat{S}}, \mathbf{x}^\mathbf{\hat{S}} ) = \Phi (\mathbf{S}) ,   \;\;\;\;   (\mathbf{\hat{T}}, \mathbf{x}^\mathbf{\hat{T}} ) = \Phi (\mathbf{T})
$$
Where $\mathbf{\hat{S}} \in \mathbb{R}^{\hat{n}\times3}$ and  $\mathbf{\hat{T}} \in \mathbb{R}^{\hat{m}\times3}$ are the output point clouds, $\mathbf{x}^\mathbf{\hat{S}} \in \mathbb{R}^{\hat{n}\times d}$ and $\mathbf{x}^\mathbf{\hat{T}} \in \mathbb{R}^{\hat{m}\times d}$ are the extracted features with dimension $d=528$.

We build $\Phi$ based on the  KPFCN backbone~\cite{thomas2019kpconv}.
KPFCN possesses the inductive bias of translation equivariance and locality, which are well suited for local geometry feature extraction.
The default KPFCN has an UNet-like structure with the same number of pooling/un-pooling layers in the encoder/decoder. We remove the decoder units after the 2nd to the last un-pooling layer from the KPFCN backbone. Thus the output $\mathbf{\hat{S}}$ and $\mathbf{\hat{T}}$ are the down-sampled point clouds. See supplementary for details.
This down-sampling is crucial for efficient computation in the following transformer (Sec.~\ref{sec:transformer}) and matching ((Sec.~\ref{sec:matching}) layers, where the time complexity are both $O(n^2)$ of the number of points.

\begin{figure*}[!t]
    \centering
\includegraphics[width=1\linewidth]{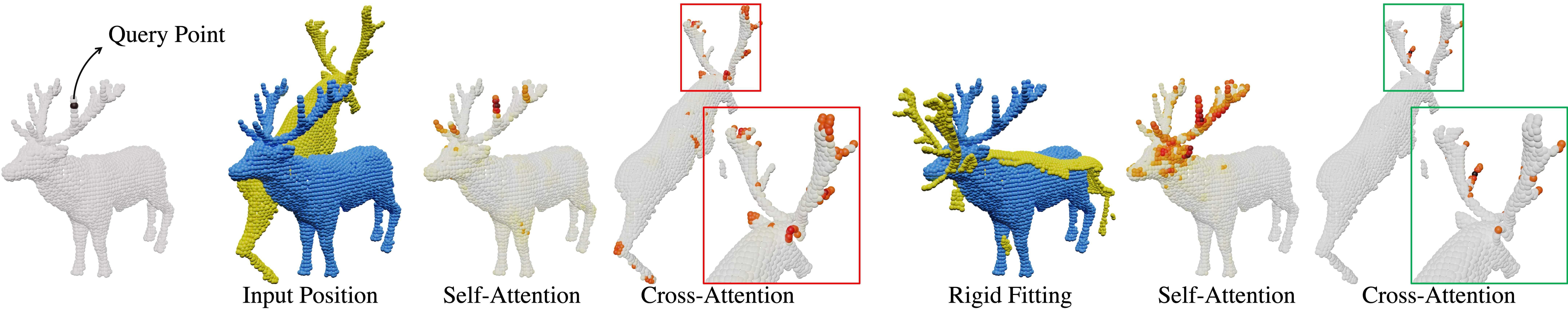}
\caption{
Visualization of self/cross attention heat maps and the rigid fitting based repositioning.  In the 2nd TMP layer, self-attention expands to cover larger context, and cross attention converges to the corresponding region. 
}
\label{fig:attention4d}
\end{figure*}

\subsection{Relative 3D Positional Encoding}
\label{sec:3d_pe}
The KPFCN backbone learns strict translation invariant features.
Translation invariant could cause ambiguity in scenes that have symmetric structures or globally repetitive geometry patches.
To resolve such ambiguity, we enhance features with the transformation sensitive 3D positional information.

We use the Rotary positional encoding proposed in~\cite{su2021roformer} and extend it to the 3D case.
Given a 3D point $\mathbf{S}_i=(x,y,z)\in \mathbb{R}^3$, 
and the it's feature $\mathbf{x}^{\mathbf{S}}_i\in \mathbb{R}^d$.
The position encoding function $\mathcal{PE}: \mathbb{R}^3 \times \mathbb{R}^d \mapsto  \mathbb{R}^d$ is defined by

\begin{center}
$
\mathcal{PE}(\mathbf{S}_i, \mathbf{x}^\mathbf{ S}_i)=
\Theta(\mathbf{S}_i)  \mathbf{x}^\mathbf{ S}_i=
$
\scalebox{0.65}{
$
\begin{pmatrix}
M_1 & & & \\
& M_2 & & \\
& & \ddots & \\
& & & M_{d/6}
\end{pmatrix}
$
}    
$ \mathbf{x}^\mathbf{ S}_i$
\end{center}
where $\Theta(\mathbf{S}_i)$ is a block diagonal matrix. Each diagonal block with size $6\times 6$ is defined by 
\begin{center}
\scalebox{0.7}{
$
M_k=
\begin{pmatrix}
\cos x\theta_k  & -\sin x\theta_k  &0&0&0&0 \\
\sin x\theta_k  & \cos x\theta_k  &0&0&0&0 \\
0&0&\cos y\theta_k  & -\sin y\theta_k  &0&0\\
0&0&\sin y\theta_k  & \cos y\theta_k  &0&0 \\
0&0&0&0&\cos z\theta_k  & -\sin z\theta_k  \\
0&0&0&0&\sin z\theta_k  & \cos z\theta_k \\
\end{pmatrix} $
}    
\end{center}
where $\theta_k = \frac{1}{10000^{6(k-1)/d}}$, 
\scalebox{0.7}{$k\in [1,2,..,d/6]$} encodes the index in the feature channel.

Compared to the Sinusoidal encoding~\cite{attention_is_all_u_need,carion2020deter}, it has two advantages: 1) $\Theta(\cdot)$ is an orthogonal function, the encoding only changes the feature's direction but not the feature's length, which could potentially stabilize the learning process. 2) The dot product of two encoded features $ \langle \mathcal{PE}(\mathbf{S}_i,  \mathbf{x}^\mathbf{ S}_i), \mathcal{PE}(\mathbf{S}_j,  \mathbf{x}^\mathbf{ S}_j)  \rangle$ can be derived to:
\begin{equation}\label{eqn:relative_position}
[ 
\Theta(\mathbf{S}_i)
\mathbf{x}^\mathbf{ S}_i  
] ^T 
\Theta(\mathbf{S}_j) 
\mathbf{x}^\mathbf{ S}_j 
= (\mathbf{x}^\mathbf{ S}_i)^T \Theta(\mathbf{S}_j-\mathbf{S}_i)  \mathbf{x}^\mathbf{ S}_j
\end{equation}
which means the relative 3D distance information can be explicitly revealed by the dot product.
We adopt this positional encoding in both the transformer (Sec.~\ref{sec:transformer}) and matching (Sec.~\ref{sec:matching}) layers.
The comparison with Sinusoidal encoding can be found in Sec.~\ref{sec:ablation}.

\subsection{Transformer}\label{sec:transformer}

After the local geometry extraction, $\mathbf{x}^\mathbf{\hat{S}}$ and $\mathbf{x}^\mathbf{\hat{T}}$ are passed through the transformer block with a self attention layer to aggregate the global context, followed by a cross attention layer to exchange information between two point clouds.  Following~\cite{attention_is_all_u_need}, the attention operation selects the relevant information by measuring the similarity between the query vector $\mathbf{q}$, and the key vector $\mathbf{k}$. 
The output vector is the sum of the value vector $\mathbf{v}$ weighted by the similarity scores. 

\medskip
\noindent
\textbf{Self Attention Layer.}
In self attention layer, $\mathbf{q}$ and ($\mathbf{k},\mathbf{ v}$) are obtained from the same point clouds (either from source or the target).  Below shows the example of self attention for the source $\mathbf{\hat{S}}$.
The vectors $\mathbf{q},\mathbf{k},\mathbf{ v}$ are first computed by
\begin{equation}\label{eqn:attention}
    \resizebox{0.9\hsize}{!}{%
    $
\mathbf{q}_i =\Theta(\mathbf{\hat{S}}_i)  W_\mathbf{q} \mathbf{x}_i^\mathbf{\hat{S}} \;\;\;\;
\mathbf{k}_j =\Theta(\mathbf{\hat{S}}_j)  W_\mathbf{k} \mathbf{x}_j^\mathbf{\hat{S}} \;\;\;\;
\mathbf{v}_j = W_\mathbf{v} \mathbf{x}_j^\mathbf{\hat{S}} 
$
        }
\end{equation}
where $W_\mathbf{q},W_\mathbf{k},W_\mathbf{v} \in \mathbb{R}^{d\times d}$ are learnable projection matrices. The feature $ \mathbf{x}_i^\mathbf{\hat{S}}$ is finally updated by
\begin{equation}\label{eqn:transormer_output}
\mathbf{x}_i^\mathbf{\hat{S}} \gets  \mathbf{x}_i^\mathbf{\hat{S}} + \text{MLP}( \text{cat}[\mathbf{q}_i,  \sum_j a_{ij}\mathbf{v}_j ] )    
\end{equation}
where $a_{ij}=\text{softmax}(\mathbf{q}_i\mathbf{k}_j^T /\sqrt{d}) $ is the attention weight, $\text{MLP}(\cdot)$ denotes a 3-layer fully connected network, and $\text{cat}[\cdot,\cdot]$ is the concatenation operator.

\medskip
\noindent
\textbf{Cross-Attention Layer.}
In cross-attention layer, the input vectors $\mathbf{q}$ and ($\mathbf{k},\mathbf{ v}$) are obtained from different point clouds depending on the direction of cross-attention ($\mathbf{\hat{S}}\to\mathbf{\hat{T}}$ or $\mathbf{\hat{T}}\to \mathbf{\hat{S}}$ ). After replacing the contents for $\mathbf{q}$, $\mathbf{k}$, and $\mathbf{ v}$, the formations are the same with self attention.

\subsection{Position Aware Feature Matching}\label{sec:matching}

After the transformer layer, we compute the scoring matrix $\mathcal{S}$ between two point clouds as 
\begin{equation}\label{eqn:matching}
\mathcal{S}(i,j)=\frac{1}{\sqrt{d}} \langle \Theta(\mathbf{\hat{S}}_i) W_{\mathbf{\hat{S}}}\mathbf{x}_i^\mathbf{\hat{S}}, \Theta(\mathbf{\hat{T}}_j) W_{\mathbf{\hat{T}}} \mathbf{x}_j^\mathbf{\hat{T}} \rangle 
\end{equation}
where $W_{\mathbf{\hat{S}}},W_{\mathbf{ \hat{ T}} }\in\mathbb{R}^{d\times d}$ are learnable projection matrices.
The features are position-encoded such that the matching algorithm could take the spatial distance into account.
We apply softmax on both dimensions (kown as the dual-softmax operation~\cite{sun2021loftr,rocco2018neighbourhood}) to convert the scroing matrix to confidence matrix $\mathcal{C}$.
$$
\mathcal{C}(i,j) = \text{Softmax}(\mathcal{S}(i,\cdot))\cdot \text{Softmax}(\mathcal{S}(\cdot,j))    
$$
Another option for matching is the sinkhorn optimal transport algorthm as in~\cite{sarlin2020superglue}. The  comparison can be seen in Sec.~\ref{sec:ablation}.

\medskip
\noindent
\textbf{Match Prediction.}
Based on the confidence matrix $\mathcal{C}$, we select matches with confidence higher than a threshold of $\theta_c$, and further enforce with mutual nearest neighbor (MNN) criteria. The influence of $\theta_c$ can be found in the supplemental ablation study.

\subsection{Disentanglement between Position and Feature.}
As shown in Fig.~\ref{fig:pipeline}, we hold the position code and geometry features in separated data streams. They are combined only when a similarity matrix needs to be computed.  In the transformer layers  (c.f. Eqn.~\ref{eqn:attention} and ~\ref{eqn:transormer_output}), position code $\Theta(\cdot)$ is only multiplied to query $\mathbf{q}$ and key $\mathbf{k}$ but not to the value $\mathbf{v}$, i.e. $\Theta(\cdot) $ can influence the attention weights but can not be a part of the feature.  
This technique leads to the \textit{Disentanglement} between position and feature. 
Opposed to this is regarded as \textit{Entangled}, which does not hold separate data streams for position and feature, i.e. they are mixed at the very beginning of the transformer.
The comparison is seen in Sec.~\ref{sec:ablation}.

\subsection{Rigid Fitting With Soft Prucrustes}\label{sec:prucrustes}
Given the the confidence matrix $\mathcal{C}$, we select the matches set $\mathcal{K}_{soft}$ with top $\hat{n}$ matching scores, and then fit them with a rigid rotation $\mathbf{R} \in SO3$ and translation $\mathbf{t}\in \mathbb{R}^3$ .  Here $\hat{n}$ is the number of points in the source cloud $\mathbf{\hat{S}}$.
Following~\cite{arun1987procrustes}, the rotation is computed from the SVD decomposition of the matrix $H = U\Sigma V^T$.  $H\in\mathbb{R}^{3\times3}$ is obtained with:  
$$H=\sum_{(i,j)\in \mathcal{K}_{soft} } \mathcal{\tilde{C}}(i,j) \mathbf{\hat{S}}_i \mathbf{\hat{T}}_j^T $$
where $\mathcal{\tilde{C}}(i,j)$ is the normalized confidence score. Then the rotation and is computed by
$$\mathbf{R}=U\text{diag}(1,1,\text{det}(UV^T))V $$ 
Then the translation is obtained with 
$$
\mathbf{t} = 
\frac{1}{|\mathcal{K}_{soft}|} 
(\sum_{(i,\cdot)\in \mathcal{K}_{soft}}\mathbf{\hat{S}}_i 
- \mathbf{R}
\sum_{(\cdot,j)\in \mathcal{K}_{soft}}
\mathbf{\hat{T}}_j) 
$$

\subsection{Repositioning}\label{sec:Repositioning}
The position encoding in Eqn.~\ref{eqn:relative_position} can reveal 3D relative distance information between a pair of points.
However, this distance knowledge is often incorrect for point pairs from two unaligned point clouds.
In other words, this position encoding benefits self-attention but may become irrelevant or noisy signals for cross-attention and matching.
To this end, we adjust the position code $\Theta(\mathbf{\hat{S}}_i)$ of $\mathbf{\hat{S}}$ with the rigid fitting $\mathbf{R},\mathbf{t}$ from the soft Procrustes layer by
$$
\Theta(\mathbf{\hat{S}}_i) \gets  \Theta( \mathbf{R}\mathbf{\hat{S}}_i + \mathbf{t})
$$We call it as repositioning. An example is shown in Fig.~\ref{fig:attention4d}.
Intuitively, repositioning pushs corresponding points closer in the position space such that it better informs cross attention and also facilitates position aware matching.

\subsection{Supervision}\label{sec:Supervision}

\medskip
\noindent
\textbf{Matching Loss.}
We minimize the focal loss over the confidence matrix $\mathcal{C}$ returned by the matching layer. 
It is defined as:
$$
\mathcal{L}_m = -  \frac{1}{|\mathcal{K}_{gt}|} \sum_{(i,j)\in \mathcal{K}_{gt} } \alpha (1-\mathcal{C}(i,j))^{\gamma} \log \mathcal{C}(i,j)
$$
where $\alpha=0.25$ and $\gamma=2$ are empirically decided focal loss parameters as in~\cite{focal_loss}, and $\mathcal{K}_{gt}$ is the set of ground-truth matches.
During training, we warp $\mathbf{\hat{S}} $ to $\mathbf{\hat{T}} $ using the ground-truth wrap function $\mathcal{W}_{gt}$, and then collect the set of mutual nearest neighbors of the two-point clouds that are below a distance threshold as $\mathcal{K}_{gt}$.

\medskip
\noindent
\textbf{Warping Loss.}
We minimize the $L_1$ loss for the point clouds that is warped by the $\mathbf{R}$,$\mathbf{t}$ from the Procrustes layer. It is defined as
$$\mathcal{L}_w = \frac{1}{|\mathcal{O}^{ \mathbf{\hat{T}}}_{ \mathbf{\hat{S}} }|} 
\sum_{i \in \mathcal{O}^{ \mathbf{\hat{T}} }_{ \mathbf{\hat{S}}}} 
| \mathcal{W}_{gt}(\mathbf{\hat{S}}_i) - \mathbf{R}\mathbf{\hat{S}}_i - \mathbf{t} | $$
where $\mathcal{O}^{\mathbf{\hat{T}}}_{ \mathbf{\hat{S}} }$ is the set of overlapping points in $\mathbf{\hat{S}}$, and $\mathcal{W}_{gt}(\cdot)$ is the ground truth warp function.
Intuitively, in rigid cases, $\mathcal{L}_w$ regularizes the optimization by suppressing false-positives in $\mathcal{K}_{soft}$ and also encourages sub-point accuracy for soft correspondences; 
in deformable cases, $\mathcal{L}_w$ tries to approximate a ``root'' pose that aligns the principal part of the overlapping region, e.g. the deer in Fig.~\ref{fig:attention4d}.

\medskip
\noindent
\textbf{Total Loss.}
As shown in Fig.~\ref{fig:pipeline}, the \textbf{T}ransform-\textbf{M}atching-\textbf{P}rocrustes (TMP) block repeats two times. The total loss combines the matching loss and warpping loss from the 1st and 2nd TMP blocks:
$
\mathcal{L} = (\mathcal{L}_m^1 + \mathcal{L}_m^2)  + \lambda_w (\mathcal{L}_w^1 + \mathcal{L}_w^2 )
$
where $\lambda_w$ is the weighting factor of warpping loss.  We show ablation study for $\lambda_w$ and the number of TMP blocks (2, 3, and 4) in supplementary.

\begin{figure}[!b]
    \centering
    \includegraphics[width=1\linewidth]{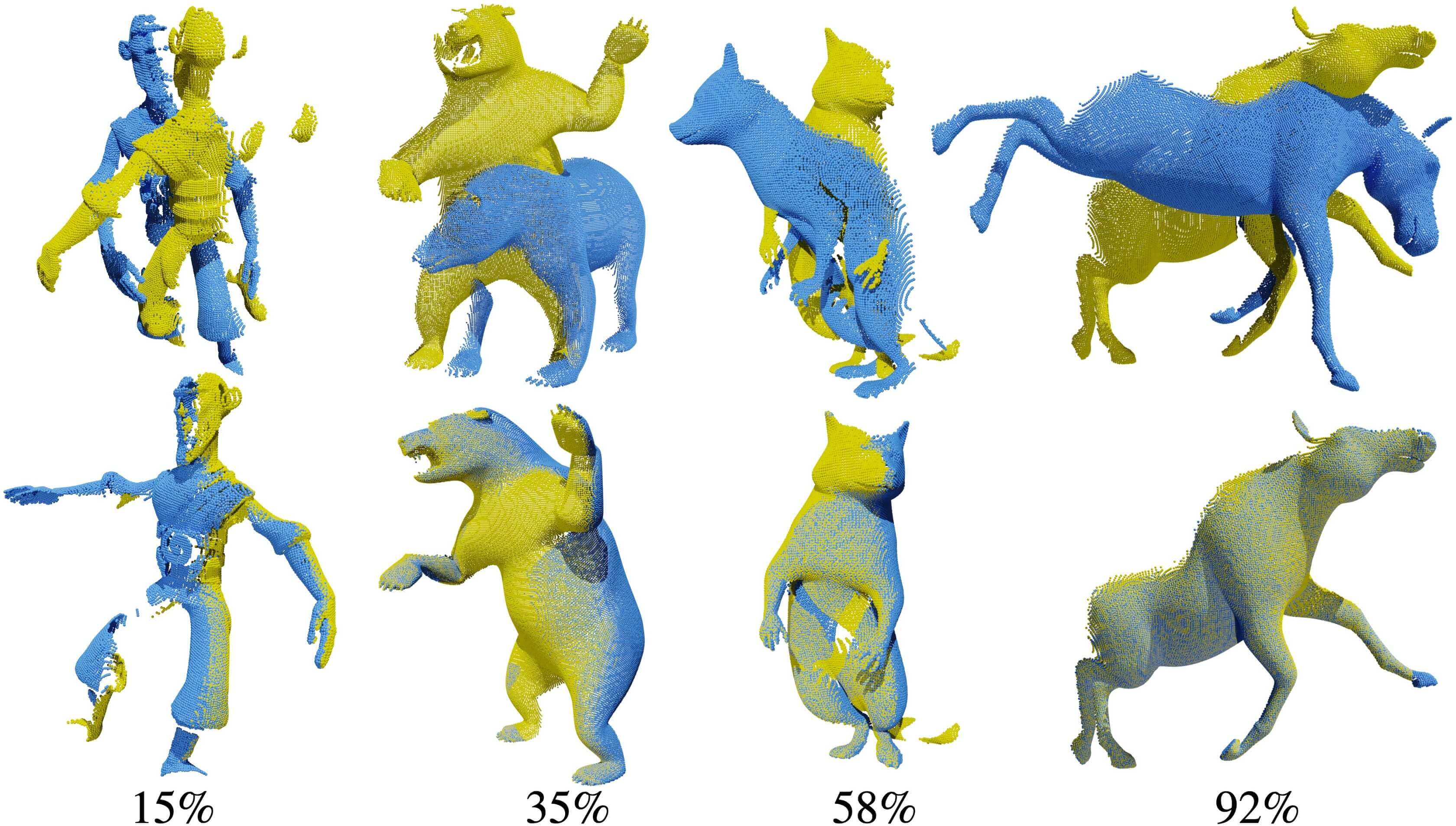}
    \caption{Examples in 4DMatch/4DLoMatch with different overlap ratio. Overlap ratio is computed relative to the source (Blue). Partial overlap is the joint effect of scene deformation and camera viewpoint change.
    }
    \label{fig:4dmatch_eg}
\end{figure}

\begin{figure}[!b]
    \centering
    \includegraphics[width=1\linewidth]{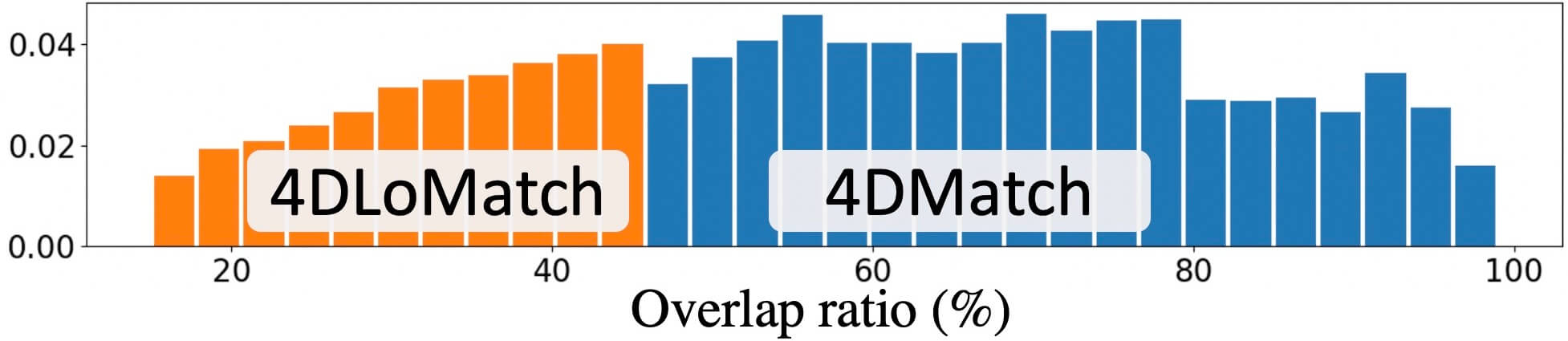}
    \caption{Histogram of the 4DMatch and 4DLoMatch benchmark. The overlap ratio threshold is set to 45\%.}
    \label{fig:statistic_4dmatch}
\end{figure}

\begin{figure*}[!t]
\centering
\includegraphics[width=1\linewidth]{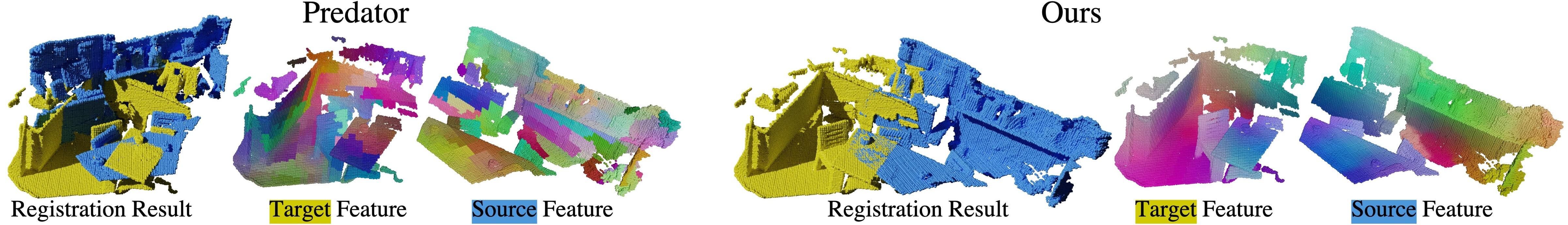}

\caption{
\textbf{Qualitative rigid point cloud registration results in 3DLoMatch benchmark.}
We use T-SNE to reduce the feature dimension to 3 and then normalize it to [0, 255] as RGB values. 
We propagate our output feature $\Theta(\mathbf{\hat{S}}_i) W_{\mathbf{\hat{S}}}\mathbf{x}_i^\mathbf{\hat{S}}$ and $ \Theta(\mathbf{\hat{T}}_j) W_{\mathbf{\hat{T}}} \mathbf{x}_j^\mathbf{\hat{T}}$ as in Eqn.~\ref{eqn:matching} to the raw point clouds via interpolation as Eqn.~\ref{eqn:inverse_interpolation}.
The feature visualization shows that our approach learns a position-aware feature representation and also accurately reflects the inter-fragment relative position.
}
\label{fig:3dmatch_compare}

\end{figure*}

\begin{figure*}[!t]
    \centering
     
\includegraphics[width=1\linewidth]{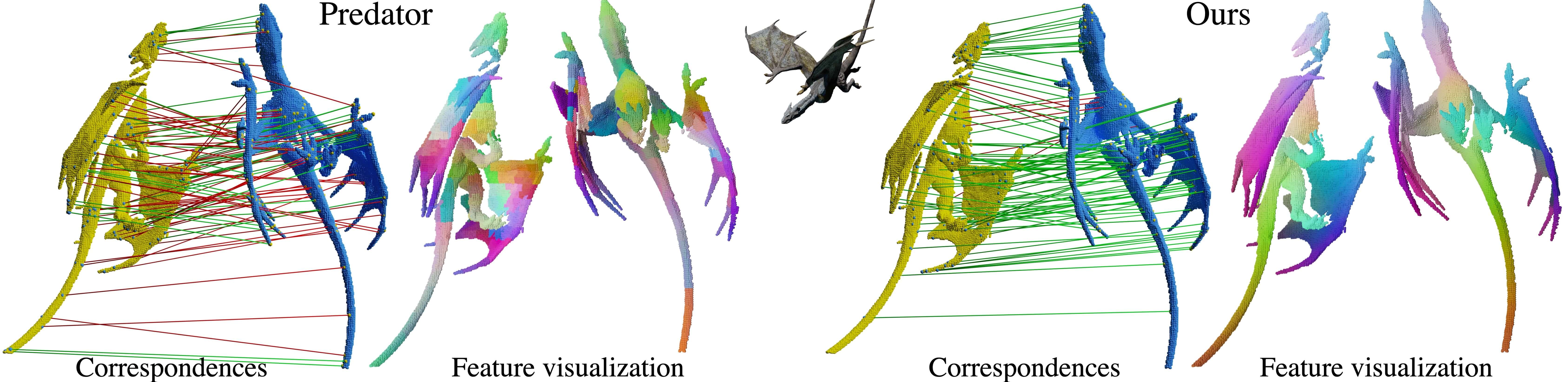}

\caption{
\textbf{Qualitative deformable point cloud matching results in 4DMatch benchmark.}
Blue/red lines indicate inliers/outliers. 
In this example, the ``Flying Dragon" has a bilateral symmetric shape.
T-SNE feature visualization shows that Predator~\cite{huang2021predator} learns similar features for the two wings of the dragon, while our method can discriminate between the left and right wings.
}
\label{fig:4dmatch_compare}

\end{figure*}

\section{4DMatch}

We propose 4DMatch, a benchmark for matching and registration of partial point clouds with time-varying geometry.
4DMatch is constructed using the sequence from DeformingThings4D~\cite{li20214dcomplete} which contains 1,972 animation sequences with ground truth dense correspondence.
We randomly select 1761 animations and generate partial point cloud scans by synthesizing depth images.
The selected 1761 sequences are divided into 1232/176/353 as train/valid/test sets.
Point cloud pairs in the 353 testing sequence are eventually split into either 4DMatch or 4DLoMatch based on an overlap ratio threshold of 45\%.
Fig.~\ref{fig:4dmatch_eg} shows examples of point cloud pairs with different overlap ratios.
Fig.~\ref{fig:statistic_4dmatch} is the histogram of overlap ratio.
The following shows the evaluation metrics for 4DMatch.

\smallskip
\noindent
\textbf{Inlier ratio (IR).} 
IR measures the fraction of correct matches in the predicted correspondences set $ \mathcal{K}_{pred} $.
A match is correct if it lies within a threshold  $\sigma=0.04\text{m}$ after the transformation using the ground truth warp function $\mathcal{W}_{gt}$.
It is defined as
\begin{equation}\label{eqn:ir}
\text{IR} = \frac{1}{\left|\mathcal{K}_{pred}\right|} \sum_{(\mathbf{p},\mathbf{q})\in \mathcal{K}_{pred}} 
\big[ ||\mathcal{W}_{gt}(\mathbf{p}) - \mathbf{q}||_2<\sigma \big]
\end{equation}
where $||\cdot||_2$ is the Euclidean norm and $[\cdot]$ is the Iverson bracket.

\smallskip
\noindent
\textbf{Non-rigid Feature Matching Recall (NFMR).}
NFMR measures the fraction of ground-truth matches $ (\mathbf{u}\in\mathbb{R}^3,\mathbf{v}\in\mathbb{R}^3) \in \mathcal{K}_{gt}$ that can be successfully recovered by the predicted correspondences $(\mathbf{p}\in\mathbb{R}^3,\mathbf{q}\in\mathbb{R}^3)\in \mathcal{K}_{pred}$.
Based on $\mathcal{K}_{pred}$, a sparse 3D scene flow filed $ \mathcal{F} = \{ \mathbf{q} - \mathbf{p} | (\mathbf{p},\mathbf{q}) \in \mathcal{K}_{pred}  \} $ for the set of anchor points  $\mathcal{A}=\{ \mathbf{p} | (\mathbf{p},\mathbf{q}) \in \mathcal{K}_{pred}\}$ are constructed.
Then the flow field $\mathcal{F}$ is propagated from $\mathcal{A}$ to a source point $\mathbf{u}$ in $ \mathcal{K}_{gt}$  using the inverse distance interpolation
\begin{equation}\label{eqn:inverse_interpolation}
\mathbf{\Gamma} (\mathbf{u}, \mathcal{A},\mathcal{F})= \sum_{\mathcal{A}_i \in \text{knn} ( \mathbf{u}, \mathcal{A})  }  \frac{
\mathcal{F}_i ||\mathbf{p} - \mathcal{A}_i||_2^{-1} 
}{\sum_{\mathcal{A}_i \in \text{knn} ( \mathbf{u}, \mathcal{A})   }
||\mathbf{u} - \mathcal{A}_i||_2^{-1}
}
\end{equation}
where $\text{knn}(\cdot,\cdot)$ denotes the $k$-nearest neighbors search with $k=3$. 
Finally, we define NFMR as
$$\text{NFMR} = \frac{1}{\left|\mathcal{K}_{gt}\right|}  \sum_{(\mathbf{u},\mathbf{v})\in \mathcal{K}_{gt}} 
\big[ ||  \mathbf{\Gamma} (\mathbf{u}, \mathcal{A},\mathcal{F}) - \mathbf{v}||_2<\sigma \big] $$Note that NFMR directly measures the fraction of ground-truth correspondences that are ``recalled". It is a different concept from the Feature Matching Recall (FMR) in the rigid case~\cite{3dmatch}.

\section{Experimental Results}

\smallskip
\noindent
\textbf{3DMatch and 3DLoMatch.}
3DMatch~\cite{3dmatch} and 3DLoMatch~\cite{huang2021predator} are a benchmark of indoor rigid scan matching and registration.
3DMatch contains scan pairs with overlap ratios greater than 30\%, while 3DLoMatch contains scan pairs with overlap ratios between 10\% and 30\%.
Following previous works~\cite{huang2021predator,yu2021cofinet}, we report the following metrics: Inlier Ratio (IR), Feature Matching Recall (FMR), rigid Registration Recall (RR), Relative Rotation Error (RTE), and Relative Translation Error (RTE). 
RR is widely recognized as the ultimate metric because it measures the fraction of successfully registered scan pairs.

\smallskip
\noindent
\textbf{Impementation details.}
Our method is implemented using Pytorch and trained using SGD on Nvidia A100 (80G) GPU.
We use a batch size of 8 and apply padding and masking to handle different point clouds sizes.
On 3DMatch, we follow the train/validation split as Predator~\cite{huang2021predator}.
The training on 3DMatch and 4DMatch both converge around the 15th epoch. We save the model with the best validation loss. 
More implementation details are seen in the supplementary.

\begin{table}[!b]

\newcolumntype{L}{>{\centering\arraybackslash}m{2cm}}
\centering
\resizebox{1\columnwidth}{!}{
\begin{tabular}{ll|cc|cc}
\toprule
                            && \multicolumn{2}{c|}{4DMatch} & \multicolumn{2}{c}{4DLoMatch} \\
\multicolumn{2}{l|}{Ablation Target}  & NFMR$\uparrow$   & IR$\uparrow$  & NFMR$\uparrow$ &IR$\uparrow$ \\ 

\midrule
 
\multirow{2}{*}{Fea. \& Pos. }  & Entangled  &\textbf{84.1}&\textbf{83.5}&63.3&51.4\\
                              &  Disentangled\textbf{*}  & 83.7& 82.7 &\textbf{66.9}  &\textbf{55.7}\\
\midrule
\multirow{2}{*}{PE type }  & Absolute &\textbf{83.7} &82.2 &63.1&51.8\\
        &  Relative\textbf{*}    & \textbf{83.7}& \textbf{82.7} &\textbf{66.9}  &\textbf{55.7}\\
        
\midrule

                        & Random rotation   &79.2 & 78.2 &58.4&46.7\\
                        & w/o Repositioning   &80.8 & 80.5 &63.6&53.7\\
Positioning             & Repositioning\textbf{*} &\textbf{83.7}& \textbf{82.7} &\textbf{66.9}  &\textbf{55.7}\\
                        & Oracle rigid fitting   &\gray{91.5}&\gray{89.7}&\gray{80.2}&\gray{67.2}\\
                        & Oracle deformation   &\gray{100.0}&\gray{99.8}&\gray{100.0}&\gray{97.6}\\

\midrule

\multirow{2}{*}{Matching }  & Sinkhorn &81.7&77.4&59.6&46.1\\
 &  Dual-Softmax\textbf{*}  & \textbf{83.7}& \textbf{82.7} &\textbf{66.9}  &\textbf{55.7}\\
\bottomrule
\end{tabular}

}
\caption{ \textbf{Ablation study on 4DMatch}. \textbf{*} indicates the default configuration of our method.
}
\label{tab:ablation_4dmatch}
\end{table}

\subsection{Ablation Study}\label{sec:ablation}

\medskip
\noindent
\textbf{--Does disentangling position and feature help?}
For non-rigid cases, disentangling position and feature achieve similar results to the entangled version on 4DMatch and achieve significantly better results on 4DLoMatch (c.f. Tab.~\ref{tab:ablation_4dmatch}).
Fea. \& Pos. disentanglement also gain $+1.1\%$ / $+2.5\%$ RR on 3DMatch / 3DLoMatch (c.f. Tab.~\ref{tab:threedmatch}). 

\medskip
\noindent
\textbf{--Positional encoding: absolute vs relative.}
Our relative 3D positional encoding yields
$+3.8\%$ higher NFMR on 4DLoMatch and
$+1.5\%$ / $+2.3\%$ higher RR on 3DMatch / 3DLoMatch
than the absolute sinusoidal encoding~\cite{attention_is_all_u_need} (c.f. Tab.~\ref{tab:ablation_4dmatch},~\ref{tab:threedmatch}).

\medskip
\noindent
\textbf{--Does Repositioning make sense?}
Repositioning gains 
$+2.9\%$ / $+3.3\%$ NFMR on 4DMatch / 4DLoMatch and 
$+0.9\%$ / $+1.0\%$ RR on 3DMatch / 3DLoMatch (c.f. Tab.~\ref{tab:ablation_4dmatch},~\ref{tab:threedmatch}).
Note that the performances drop significantly with the $\textit{Random rotation}$-based positioning. 
The \textit{Oracle deformation} in Tab.~\ref{tab:ablation_4dmatch} and \textit{Oracle rigid fitting} in Tab.~\ref{tab:threedmatch} achieve near perfect results. 
These results demonstrate the importance of positional knowledge for point cloud registration. In Tab.~\ref{tab:ablation_4dmatch}, the \textit{Oracle rigid fitting} refers to the best rigid fitting for the ground truth deformation. We consider it as the upper bound of the rigid fitting-based repositioning approach.

\smallskip
\noindent
\textbf{--Matching algorithm: Sinkhorn vs dual softmax.}
The dual softmax operator achieves higher scores on all benchmarks than the Sinkhorn approach.

\begin{figure*}[!t]
    \centering
     
\includegraphics[width=1\linewidth]{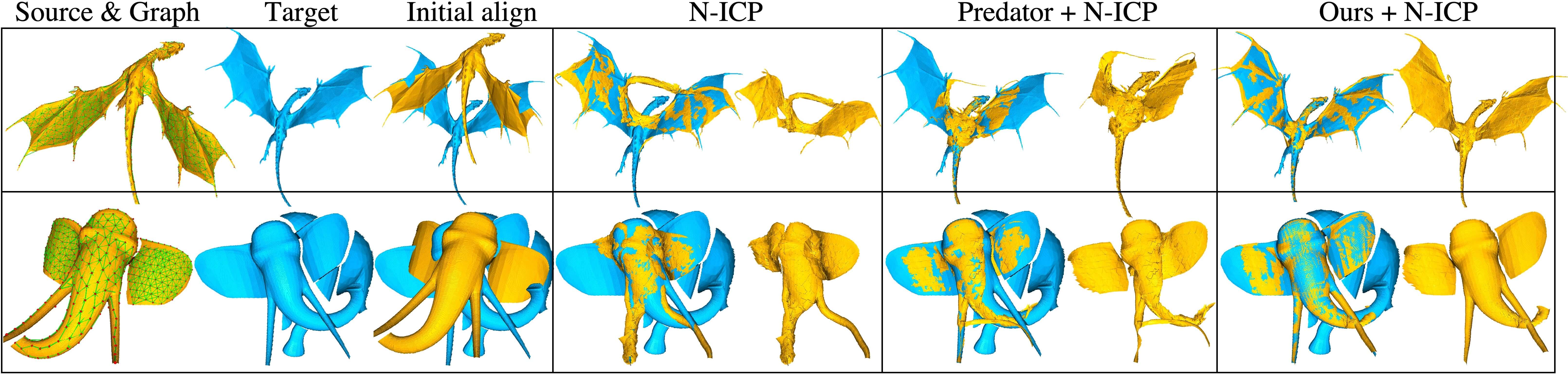}

\caption{Qualitative non-rigid registration results. As shown on the left end, scene deformation is approximated by the deformation graph.
Results shown are the final alignment and the transformed source. 
}
\label{fig:4dregistration}

\end{figure*}

\begin{table}[!t]
\centering

\resizebox{1\columnwidth}{!}{
\begin{tabular}{c|l|c|cc|cc}
\toprule
    &                      &&   \multicolumn{2}{c|}{4DMatch} & \multicolumn{2}{c}{4DLoMatch} \\
 Category &Method& S{$^\dagger$} &  NFMR$\uparrow$   & IR$\uparrow$ &  NFMR$\uparrow$&IR$\uparrow$ \\ 

\midrule
\multirow{2}{*}{\makecell{ Probablistic Model}}  
&CPD~\cite{Coherent_point_drift}   &&6.0&5.5&0.4&0.2\\  
&BCPD~\cite{bayesian_coherent_point_drift}  &&11.4&10.1&1.2& 0.5\\  

\midrule

\multirow{2}{*}{\makecell{Functional Map}{$^\ddagger$}} 
&ZoomOut~\cite{melzi2019zoomout}  & &4.2&3.8&1.3& 0.5\\  
&GeoFM~\cite{deepgeofm2020}   &$\checkmark$&25.0&20.5&11.7&4.5 \\

\midrule

\multirow{3}{*}{\makecell{Scene Flow}} 
&PWC~\cite{wu2019pointpwc}  & $\checkmark$  
& 21.6  & 20.0     &10.0   &7.2 \\  
&FLOT~\cite{puy2020flot}   &$\checkmark$
&  27.1  &  24.9      &  15.2  &  10.7  \\ 
&NSFP~\cite{Neural_Scene_Flow_Prior}   &
&18.5  & 16.3     &1.2  &0.5  \\

\midrule
\multirow{3}{*}{\makecell{ Feature Matching}}
&D3Feat~\cite{d3feat}   &$\checkmark$&55.5  &54.7    &27.4  &21.5  \\  
&Predator~\cite{huang2021predator}  &$\checkmark$&56.4  &60.4      &32.1  &27.5\\
&Ours   &$\checkmark$&\textbf{ 83.7}& \textbf{82.7}        &     \textbf{66.9}       &    \textbf{55.7}         \\
\bottomrule
\end{tabular}

}
\caption{ \textbf{Quantitative Results on 4DMatch.} S{$^\dagger$} denotes supvervised methods.
{$^\ddagger$} Point cloud Laplacian is obtained via~\cite{robust_laplacian}. }

\label{tab:4dmatch_sota}
\end{table}

\subsection{Comparison with the state-of-the-art}

\medskip
\noindent
\textbf{4DMatch \& 4DLoMatch.}
As shown in Tab.~\ref{tab:4dmatch_sota}, our method acheive $+27.1\%$ / $+34.8\%$ higher NFMR and $+22.3\%$ / $+28.2\%$ higher IR than Predator.
Fig.~\ref{fig:4dmatch_compare} shows the qualitative point cloud matching results on 4DMatch.
We also report results from probabilistic registration~\cite{Coherent_point_drift,bayesian_coherent_point_drift}, functional map-based~\cite{melzi2019zoomout,deepgeofm2020}, and scene flow methods~\cite{wu2019pointpwc,puy2020flot,Neural_Scene_Flow_Prior}.
Overall, feature matching methods~\cite{huang2021predator,d3feat} show superior performance than other categories.
Existing scene flow methods, can not handle large global motion.
Functional map methods fail because they need connected shapes to obtain reliable Laplacian matrices, while the shapes in 4DMatch are usually disconnected due to occlusion (c.f. examples in Fig.~\ref{fig:4dmatch_eg}). 
The Coherent Point Drift (CPD) models~\cite{Coherent_point_drift,bayesian_coherent_point_drift} do not work well on the partial scan data.

\medskip
\noindent
\textbf{3DMatch \& 3DLoMatch.} 
Compared to Predator~\cite{huang2021predator}, our method acheives $+1.7\%$ / $+6.6\%$ higher RR on 3DMatch / 3DLoMatch (c.f. Tab.~\ref{tab:threedmatch}).
Fig.~\ref{fig:3dmatch_compare} shows the qualitative results for a low overlap case.
As shown in Tab.~\ref{tab:rre_rte}, Our method also produces better RRE and RTE than Predator; 
the point-to-point ICP-based postprocessing can further improve the registrations.

\begin{table}[!t]
\centering
\resizebox{\columnwidth}{!}{

\begin{tabular}{l|ccc |ccc}
\toprule
                        & \multicolumn{3}{c|}{3DMatch~\cite{3dmatch}} & \multicolumn{3}{ c}{3DLoMatch~\cite{huang2021predator}} \\  
         & FMR$\uparrow$ & IR$\uparrow$   &\textbf{RR}$\uparrow$ & FMR$\uparrow$ & IR$\uparrow$   &\textbf{RR}$\uparrow$                  \\  

\midrule
3DSN~\cite{3DSN}       &94.7& 36.0   & 81.5        &61.9&11.4           & 36.6                     \\  
FCGF~\cite{FCGF}         &95.2& 56.9 & 88.2          &60.9&    21.4     & 45.8                    \\  
D3Feat~\cite{d3feat}      & 95.8 & 39.0  & 85.8          &69.3&   13.2      & 40.2                    \\  
Predator~\cite{huang2021predator} &96.7 & 58.0   & 91.8         &78.6&  26.7 &62.4                    \\  

\midrule
Ours-Entangled   &97.8&46.8&92.5&83.8&23.3&66.5   \\  
Ours-Absolute    & 97.4 & 62.0 & 92.1 &84.1&29.3&67.7   \\  
Ours-Random rotation    &97.5           &60.6          &90.8           &79.2       &25.6              & 60.3                     \\  
Ours-w/o repositioning      & 98.0 &57.6 &92.8  &84.2 &27.8  &68.0                     \\ 
Ours-Sinkhorn    &97.6&46.5&92.2  &81.7&17.7&  64.9  \\  
Ours (Default)   & 98.3 &55.5 &\textbf{93.5}  &84.5 &26.0  &\textbf{69.0}                     \\ 
\midrule
\gray{Oracle Rigid fitting}   &\gray{100.0}& \gray{99.3}  & \gray{100.0}     & \gray{100.0} &   \gray{95.0}  & \gray{99.8}                   \\ 
\bottomrule
\end{tabular}

}
\caption{\textbf{Feature matching and RANSAC registration results on 3DMatch.}
Refer to Tab.~\ref{tab:ablation_4dmatch} for the configs of Ours (Default).
Note that, in this paper, \textbf{RR} is averaged on all scan pairs to reflect the true percentage of successful registrations.
}
\label{tab:threedmatch}
\end{table}

\begin{table}[!h]
\centering
\resizebox{1.05\columnwidth}{!}{

\begin{tabular}{l|c|ccc |ccc}
\toprule
         &                & \multicolumn{3}{c|}{3DMatch } & \multicolumn{3}{ c}{3DLoMatch } \\  
          && RRE   $\downarrow$ & RTE   $\downarrow$ & \textbf{RR}  $\uparrow$  & RRE   $\downarrow$ & RTE   $\downarrow$ &\textbf{RR}  $\uparrow$\\  

\midrule 
Predator  & 
\multirow{2}{*}{RANSAC}  
& 2.72  & 7.8  & 91.8 &  4.44 &     11.6    &62.4                   \\  
Ours  &    & \underline{2.48} &\underline{7.2} & \underline{93.5}  &\underline{4.10}& \underline{10.8}&\underline{69.0}                  \\

\midrule
Predator  & 
\multirow{2}{*}{RANSAC+ICP}
&2.06&6.2&92.3&3.46 &9.8& 65.2     \\ 

Ours  &       &\textbf{1.96}&\textbf{6.0}&\textbf{93.9}&\textbf{3.17} &\textbf{8.9}&\textbf{ 71.3}                \\ 
\bottomrule
\end{tabular}

}
\caption{\textbf{RRE (\textdegree), RTE ($cm$) and RR ($\%$) on 3DMatch.}
Point-to-point ICP can further refine the transformations.
}
\label{tab:rre_rte}
\end{table}

\subsection{Integrating to non-rigid registration.}
We further integrate the predicted matches to non-rigid point cloud registration. 
For registration, we adopt the non-rigid iterative closest point (N-ICP)~\cite{li2008regist,zollhofer2014real} method which iteratively minimizes the typical energy function: 
$$\mathbf{E}_{total}(\mathcal{G}) =\lambda_c \mathbf{E}_{corr}(\mathcal{G}) +\lambda_r\mathbf{E}_{reg}(\mathcal{G})$$ 
where $\mathbf{E}_{corr}$ is the correspondence term, $\mathbf{E}_{reg}$ is the regularization term as in~\cite{arap}, and $\mathcal{G}$ is the latent deformation graph model. The correspondence term $\mathbf{E}_{corr}$ is defined by the L2 distance between corresponding points. See the supplementary for formal definitions.

Fig.~\ref{fig:4dregistration} shows the qualitative results of non-rigid registration on the dragon and elephant examples.
The \textit{N-ICP} baseline adopts the simple Euclidean space nearest neighbor search-based correspondence.
\textit{Predator + N-ICP} and \textit{Ours + N-ICP} replace the correspondence term by the predicted matches during the beginning 20 iterations and then run 45 N-ICP iterations for refinements. Our method better informs non-rigid registration than Predator.

\section{Conclusion}
\begin{figure}[!h]
    \centering
\includegraphics[width=1\linewidth]{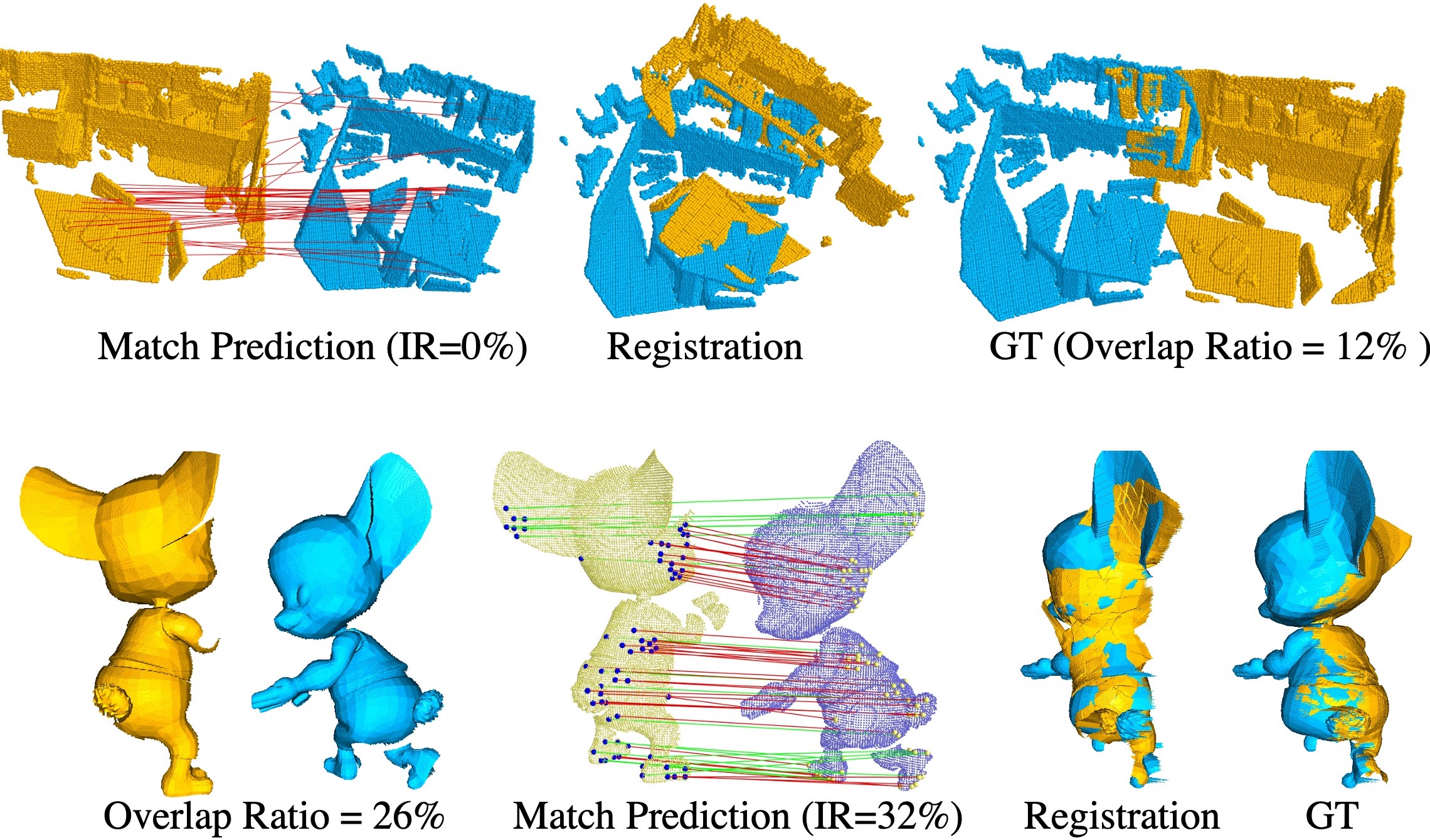} 

\caption{
\textbf{Failure cases in rigid (top) and non-rigid (bottom) matching.}
The low overlapping cases are still challenging, in particular, if the point clouds have similar patterns in the non-overlapping region. Given the similar table and chairs (top), and a similar half-body (bottom), our method fails to treat them as different features. 
}
\label{fig:failure_case}

\end{figure}

By leveraging the positional knowledge, Lepard demonstrates state-of-the-art feature matching results for both rigid and deformable point clouds.
A promising direction is extending it to end-to-end registration.  
A few limitations are yet to be addressed:
1) Lepard is a coarse matching approach. Fine-grained correspondence could be obtained with learning-based refinement, e.g.~\cite{sun2021loftr,yu2021cofinet}.
2) In deformable cases, Lepard does not explicitly handle topological changes. A potential solution is to jointly learn matching with motion segmentation.
3) Finally, matching and registration in the low overlapping cases is particularly challenging due to data incompleteness.
Fig.~\ref{fig:failure_case} shows such failure cases. A potential solution for low-overlap scenarios is to expand the mutual information via data completion~\cite{yu2021pointr,li20214dcomplete}.
Learning outlier rejection as~\cite{3DRegnet,bai2021pointdsc} could also benefit registration.

\section{Acknowledgement}
This work was partially supported by JST AIP Acceleration Research JPMJCR20U3, Moonshot R\&D Grant Number JPMJPS2011, CREST Grant Number JPMJCR2015, and Basic Research Grant (Super AI) of Institute for AI and Beyond of the University of Tokyo.

{\small
\bibliographystyle{abbrv}
\bibliography{non-rigid-reg-2021}
}

\twocolumn[{%
\begin{center}
\renewcommand{\baselinestretch}{1.5} 
\textbf{
\Large
Supplementary Material:\\
\vspace*{0.1cm}
Lepard: Learning partial point cloud matching in rigid and deformable scenes
\vspace*{0.5cm}
}
\end{center}
}]
\setcounter{equation}{0}
\setcounter{figure}{0}
\setcounter{table}{0}
\setcounter{page}{1}
\setcounter{section}{0}
\renewcommand{\thesection}{\Roman{section}}
\makeatletter
\renewcommand{\theequation}{\arabic{equation}}
\renewcommand{\thefigure}{\arabic{figure}}

\noindent 
Supplementary material includes: ablation study (Sec.~\ref{sec:s-ablation}); implementation details (Sec.~\ref{sec:imp-1},~\ref{sec:imp-2},~\ref{sec:imp-3}, and~\ref{sec:imp-4}); formal definition of non-rigid registration (Sec.~\ref{sec:s-nr}); and more results on 3DMatch and 4DMatch (Sec.~\ref{sec:s-qualitative}).

\section{Ablation study}\label{sec:s-ablation}

\medskip
\noindent
\textbf{--Influence of Warping Loss Weight.}
Applying warping loss in general yields higher NFMR and IR 4DMatch and 4DLoMatch. In particular, in the low overlap situations, the performance grows steadily with the increasing of the motion loss weight (c.f. Tab.~\ref{tab:ablation_motion_loss_4D}).
In 3DMatch and 3DLoMatch, warping loss significantly increases the Inlier Rate (IR). However, it leads to a decrease in Registration Recall (c.f. Tab.~\ref{tab:ablation_motion_loss_3D}). We assume that this is because the warping loss might suppress some border cases correspondences which could have benefited the RANSAC.
In deformable cases, a high inlier rate is desired for successful non-rigid registration.
However, in rigid cases, the inlier rate is less important since RANSAC is very robust to noise.
Therefore, we set $\lambda_w=0.1$ for 4DMatch and $\lambda_w=0.0$ for 3DMatch.

\begin{table}[!h]

\newcolumntype{L}{>{\centering\arraybackslash}m{2cm}}
\centering
\resizebox{.7\columnwidth}{!}{
\begin{tabular}{l|cc|cc}
\toprule
 & \multicolumn{2}{c|}{4DMatch} & \multicolumn{2}{c}{4DLoMatch} \\
 & NFMR$\uparrow$   & IR$\uparrow$  & NFMR$\uparrow$ &IR$\uparrow$ \\ 
\midrule
 
 $\lambda_w=0$  &82.9&82.4&62.1&52.2\\
 $\lambda_w=0.05$ &\textbf{85.3}&\textbf{83.9}&65.1& 54.5\\
 $\lambda_w=0.1$ \textbf{*} & 83.7& 82.7 &\textbf{66.9}  &\textbf{55.7}  \\\midrule
\end{tabular}
}
\caption{Influence of warping loss on 4DMatch.
}
\label{tab:ablation_motion_loss_4D}
\end{table}

\begin{table}[!ht]

\newcolumntype{L}{>{\centering\arraybackslash}m{2cm}}
\centering
\resizebox{0.8\columnwidth}{!}{

\begin{tabular}{l|ccc |ccc}
\toprule
                        & \multicolumn{3}{c|}{3DMatch~\cite{3dmatch}} & \multicolumn{3}{ c}{3DLoMatch~\cite{huang2021predator}} \\  
         & FMR$\uparrow$ & IR$\uparrow$   &\textbf{RR}$\uparrow$ & FMR$\uparrow$ & IR$\uparrow$   &\textbf{RR}$\uparrow$                  \\   
 
\midrule
 $\lambda_w=0$   & \textbf{98.0} &63.7 &\textbf{93.6}  &\textbf{85.6} &29.5  &\textbf{69.0}                     \\ 
 $\lambda_w=0.05$ & 97.8  & 66.6 & 92.9 &84.1  & 36.5  &67.9\\
 $\lambda_w=0.1$  & 97.6  & \textbf{71.5} & 92.9 &84.6  &\textbf{ 38.8}  &68.2  \\\midrule
\end{tabular}
}
\caption{Influence of warping loss on 3DMatch.
}
\label{tab:ablation_motion_loss_3D}
\end{table}

\medskip
\noindent
\textbf{--Influence of Confidence Threshold.}
In rigid cases, increasing the confidence threshold of correspondence leads to a decrease in registration recall (c.f. Tab.~\ref{tab:ablation_conf_thr_3D}).
Same to the above ablation, we assume that this is because increasing the confidence threshold inevitably suppresses some borderline correspondences which could have benefited the RANSAC. In deformable cases, increasing the confidence threshold results in a higher IR but getting a lower NFMR (c.f. Tab.~\ref{tab:theta_ablation_4d}). We found $\theta_c$=0.1 a good trade-off between precision and recall.

\begin{table}[!ht]

\newcolumntype{L}{>{\centering\arraybackslash}m{2cm}}
\centering
\resizebox{.7\columnwidth}{!}{
\begin{tabular}{l|cc}
\toprule
 &  3DMatch (RR$\uparrow$)  &  3DLoMatch (RR$\uparrow$) \\
\midrule
$\theta_c = 0.05$ & \textbf{93.6} & \textbf{69.0} \\
$\theta_c = 0.1$ & 92.3 & 67.9 \\
$\theta_c = 0.15$ & 91.7 & 67.0 \\
$\theta_c = 0.2$ & 91.2 & 65.3 \\
\bottomrule
\end{tabular}
}
\caption{ Influence of confidence thresholds on 3DMatch and 3DLoMatch.}
\label{tab:ablation_conf_thr_3D}
\end{table}

\begin{table}[!h]
\centering
\resizebox{1\columnwidth}{!}{
\begin{tabular}{l|ccc|ccc}
\toprule
                          &   \multicolumn{3}{c|}{4DMatch} & \multicolumn{3}{c}{4DLoMatch} \\
 Method & $|\mathcal{K}_{pred}|$ & NFMR$\uparrow$   & IR$\uparrow$ &
 $|\mathcal{K}_{pred}|$   & NFMR$\uparrow$&IR$\uparrow$ \\ 
\midrule
D3Feat (1000)   &267  &51.6  &52.7  & 204  &23.6  &21.2  \\  
D3Feat (3000)   &532  &55.5  &54.7  & 379  &27.4  &\underline{21.5}  \\  
D3Feat (5000)   &697  &\underline{56.1}  &\underline{55.3}  & 473  &\underline{28.1}  &21.3  \\  
\midrule
Predator (1000) &273  &53.3  &60.0  &205   &30.6  &\underline{29.8}\\
Predator (3000) &534  &56.4  &\underline{60.4}  &372   &\underline{32.1}  &27.5\\
Predator (5000) &698  &\underline{56.8}  &59.3  &480   &\underline{32.1}  &25.0\\
\midrule
Ours ($\theta_c$=0.2) &   523 &82.2   & \textbf{85.4}  & 325& 63.1  &    \textbf{60.4}       \\
Ours ($\theta_c$=0.1)\textbf{*} & 596 & 83.7& 82.7    & 407    &     66.9       &    55.7         \\
Ours ($\theta_c$=0.05) & 624 & \textbf{83.9} & 80.9    & 447  &  \textbf{67.6}       & 52.5    \\
\bottomrule
\end{tabular}

}
\caption{ Influence of confidence thresholds on 4DMatch and 4DLoMatch.
D3Feat~\cite{d3feat} and Predator~\cite{huang2021predator} probabilistically sample points either from a saliency heat map or from a machability$\times$overlap heat map (numbers in brackets are the numbers of sampled points). 
Ours uses the confidence threshold $\theta_c$ to get matches from the confidence matrix (c.f. Sec.~\ref{sec:matching}).
All methods apply the mutual nearest neighbor criteria to filter matches.
$|\mathcal{K}_{pred}|$ indicates the average number of final predicted correspondences.}
\label{tab:theta_ablation_4d}
\end{table}

\medskip
\noindent
\textbf{--Adding more TMP blocks.}
We tested 3 and 4 TMP layers. The corresponding number of the Repositioning layer is 2 and 3 because it is placed between every two consecutive TMP layers.
As shown in Tab.~\ref{tab:ablation_tmp_layer}, in 3DMatch, additional layers do not improve the results;
in 4DMatch,  3 TMP layers achieve the best results. 
Adding layers inevitably increase the training time.

\begin{table}[!h]
\newcolumntype{L}{>{\centering\arraybackslash}m{2cm}}
\centering
\resizebox{0.9\columnwidth}{!}{
\begin{tabular}{cl|cccc}
\toprule
&Number of TMP layer &2&3&4\\
&Number of Repositioning layer &1&2&3\\
\midrule
\multirow{3}{*}{Rigid} 
&RR($\%$)$\uparrow$ on 3DMatch       &\textbf{93.6}&92.8 &93.0 \\
&RR($\%$)$\uparrow$ on 3DLoMatch     &\textbf{69.0} &68.2 &68.8 \\
&Training Time (hour)$\downarrow$ &\textbf{20} &25  &31  \\
\midrule
\multirow{3}{*}{Deformable}
&NFMR($\%$)$\uparrow$ on 4DMatch     &83.7& \textbf{85.9}&84.5.\\
&NFMR($\%$)$\uparrow$ on 4DLoMatch   &66.9& \textbf{68.1}&59.6\\
&Training Time (hour)$\downarrow$ &\textbf{18}&21&24\\
\bottomrule
\end{tabular}
}
\caption{Ablation study of number of TMP layers.}
\label{tab:ablation_tmp_layer}
\end{table}
\vspace{-0.3cm}

\section{Sparse $\Theta(\cdot)$ Multiplication} \label{sec:imp-1}
Taking the advantage of the sparsity of $\Theta(\cdot)$,
given a position $\mathbf{p}=(x,y,z)\in\mathbb{R}^3$ and a feature $\mathbf{x}\in\mathbb{R}^d$,
the multiplication $\Theta(\mathbf{p}) \mathbf{x}$ can be efficiently realized by
\begin{center}
\scalebox{0.8}{
$
\begin{pmatrix}
\mathbf{x}(0) \\ \mathbf{x}(1) \\ \mathbf{x}(2) \\ \mathbf{x}(3) \\
\mathbf{x}(4) \\ \mathbf{x}(5) \\\vdots \\ 
\mathbf{x}({d/6-1}) \\ \mathbf{x}({d/6-1}) 
\end{pmatrix}
\otimes
\begin{pmatrix}
\cos x\theta_0 \\ \cos x\theta_0 \\
\cos y\theta_0 \\ \cos y\theta_0 \\
\cos z\theta_0 \\ \cos z\theta_0 \\
 \vdots \\ \cos z\theta_{d/6 -1} \\ \cos z\theta_{d/6 -1} 
\end{pmatrix} + 
\begin{pmatrix}-\mathbf{x}(1) \\ \mathbf{x}(0) \\ -\mathbf{x}(3) \\ \mathbf{x}(2) \\-\mathbf{x}(5) \\ \mathbf{x}(4) \\
\vdots \\ -\mathbf{x}({d/6-1}) \\ \mathbf{x}({d/6-1}) 
\end{pmatrix}\otimes
\begin{pmatrix}
\sin x\theta_0 \\ \sin x\theta_0 \\ 
\sin y\theta_0 \\ \sin y\theta_0 \\ 
\sin z\theta_0 \\ \sin z\theta_0 \\ 
\vdots \\ 
\sin z\theta_{d/6-1} \\ \sin z\theta_{d/6-1} 
\end{pmatrix}$
}    
\end{center}

\section{Hyper Parameters}\label{sec:imp-2}
\begin{table}[!ht]
    \centering
    \begin{threeparttable}
    \resizebox{.99\columnwidth}{!}{
    \begin{tabular}{l|l|c|c }
    \toprule
                        && 3DMatch       & 4DMatch     \\ \midrule
\multirow{4}{*}{Metric }    & Inlier threshold   &  0.1$m$ &  0.04$m$  \\
           & RR threshold  &  0.2$m$ &  -- \\
           & FMR threshold &  5\%    &  --  \\
           & NFMR threshold& --      & 0.04$m$ \\ 
    \midrule
\multirow{2}{*}{Match Prediction} & Confidence threshold $\theta_c$    & 0.05    & 0.1 \\
             & Apply MNN & False & True \\ 
    \midrule            
KPFCN Config & Input subsampling radius    & 0.025$m$    & 0.01$m$ \\
    \midrule
\multirow{2}{*}{Supervision}  & GT match radius & 0.06$m$ & 0.024$m$ \\ 
            & Warping loss weight $\lambda_w$ & 0.0 & 0.1 \\

    \bottomrule
    
    \end{tabular}
    
    }
    \begin{tablenotes}\footnotesize
    \item[]RR: Registration Recall
    \item[]FMR: Feature Matching Recall
    \item[]NFMR: Non-rigid Feature Matching Recall
    \item[]MNN: Mutual Nearest Neighbor
    \end{tablenotes}
    \end{threeparttable}
    \caption{The hyper parameters for metric evaluation, match prediction, KPFCN backbone, and training loss}
    \label{tab:hyperparam}
\end{table}

\section{Time and memory expense} \label{sec:imp-3}
 
\begin{table}[!ht]

\newcolumntype{L}{>{\centering\arraybackslash}m{2cm}}
\centering
\resizebox{.75\columnwidth}{!}{
\begin{tabular}{l|c|c}
\toprule
&Predator~\cite{huang2021predator}& Lepard (Ours)\\
\midrule
Average time (s) & 0.18  & \textbf{0.10} \\ 
Cuda memory (MB) & 13,361  &\textbf{6,595}  \\ 
\bottomrule
\end{tabular}
}
\caption{Time and Cuda memory usage of inference on an Nvidia A100 (80G) GPU. Time is averaged on 2193 testing samples in 4DLoMatch.
Lepard is about twice as efficient as Predator on both time and memory.
}
\label{tab:time_cost}
\end{table}

\begin{table}[!ht]

  \centering
	\resizebox{1\linewidth}{!}{
    \begin{tabular}{ccccc}
	\toprule
    KPFCN & Self Att. ($\times2$) &  Cross Att. ($\times2$) &  Matching ($\times2$) &  Procrustes ($\times2$)  
    \\\midrule    
    0.0109 &
    0.0016 ($\times2$)&
    0.0014 ($\times2$)&
    0.0023 ($\times2$)&
    0.0191 ($\times2$) 
    \\    
	\bottomrule
			
	\end{tabular}
	}
	\caption{Average time (s) of Lepard function inference on an Nvidia A100 (80G) GPU. Time is averaged on 2193 testing samples in 4DLoMatch.
	}
	\label{tab:avg_iteration_time}

\end{table}

\section{KPFCN backbone architecture} \label{sec:imp-4}

\begin{figure}[!ht]
    \centering
    \includegraphics[width=1\linewidth]{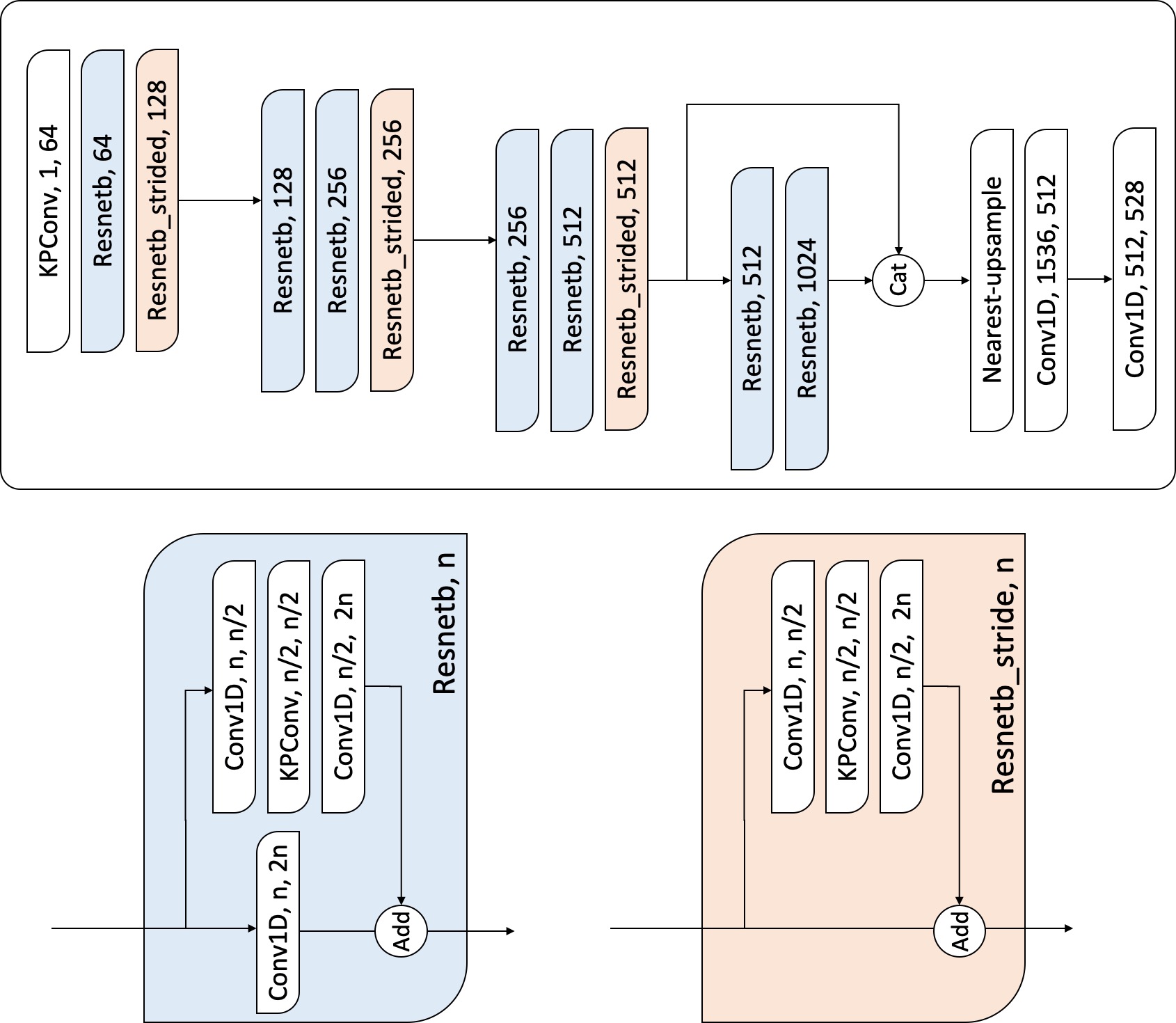}
    \caption{Details of the KPFCN backbone architecture.}
    \label{fig:backbone_detail}
\end{figure}
\section{Non-Rigid Registration}\label{sec:s-nr}
This section introduce the non-rigid registration technique used in this paper.

\begin{figure}
    \centering
    \includegraphics[width=1\linewidth]{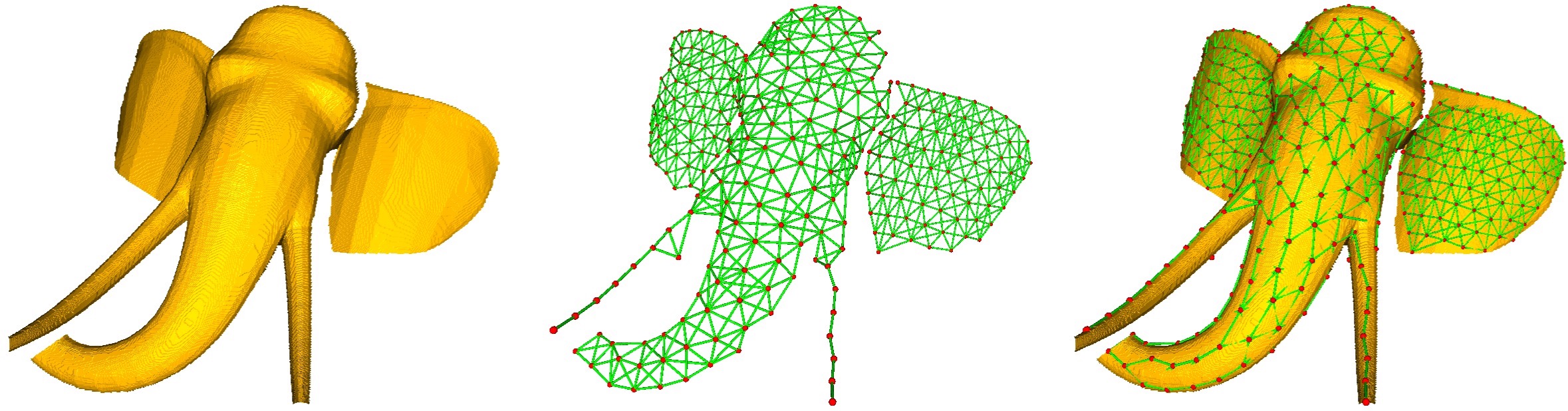}
    \caption{Deformation model. 
    Nodes $\mathcal{V}$ (red dot) are evenly sampled over the source surface. Edges $\mathcal{E}$ (green lines) are computed between nodes based on geodesic connectivity. The point cloud examples in 4DMatch are obtained from depth images. To compute geodesic distance, we construct the surface triangle mesh by connecting the nearby pixels' 3D locations. We filter the triangles with an edge larger than 4cm. During registration, for numerically stable optimization, we ignore point cloud clusters with fewer than 40 deformation nodes.
    }
    \label{fig:deformation_graph}
\end{figure}

\medskip
\noindent
\textbf{Deformation Model.}
To represent the dense motion from a source to a target, we adapt the embedded deformation model of Sumner et al.~\cite{embededdeformation}. 
The non-rigid deformation is parameterized by the deformation graph $\mathcal{G} = \{\mathcal{V}, \mathcal{E}\}$, where $\mathcal{V}$ is the set of node and $\mathcal{E}$ is the set of edge. As shown in Fig.~\ref{fig:deformation_graph}, we evenly sample graph nodes $\mathcal{V}$ over the source point cloud surface.
Each point in the scene has a 3D location: $\mathbf{g}_i \in \mathbb{R}^3 $. 
The motion of a node $i\in\mathcal{V}$ is parameterized by a translation vector:  $\mathbf{t}_i \in \mathbb{R}^3 $ and a rotation matrix: $\mathbf{R}_i \in \mathrm{SO}3 $. 
In addition, we represent rotations by $\mathbf{R}_i = \exp(\mathbf{\varphi}^{\wedge}_i)\mathbf{R}_i $, where $\mathbf{\varphi}_i= [0,0,0]$ represents the delta of the rotation in axis-angle form. $(\mathbf{\cdot}) ^\wedge$operator convert a 3-dimensional vector to a 3 × 3 skew-symmetric matrix. $\exp : \mathfrak{so3} \mapsto \mathrm{SO3}$  map the skew-symmetric matrix to 3 × 3 rotation matrix using the Rodrigues formula.
Finally, all unkowns in the graph are
$$\mathcal{G}=\left(   \mathbf{\varphi}_1 ,  \cdots, \mathbf{\varphi}_{|\mathcal{V}|} |  \mathbf{t}_1  , \cdots , \mathbf{t}_{|\mathcal{V}|}    \right) $$

\medskip
\noindent
\textbf{Non-rigid Warping Function}
Given a point $\mathbf{p} \in \mathbb{R}^3 $, the non-rigid warping function $\mathcal{W}$ is defined as
$$
\mathcal{W}(\mathbf{p}) = \sum_{i\in \mathcal{V}} w_{\mathbf{p},i} ( \mathbf{R}_i (\mathbf{p}-\mathbf{g}_i) + \mathbf{g}_i +\mathbf{t}_i ) 
$$
where $w_{\mathbf{p},i} \in \mathbb{R}$ is the ``skinning weight" that measure the influence of node $i$. They are computed as 
$$
w_{\mathbf{p},i} = Ce^{\frac{1}{2\gamma^2}|| \mathcal{V}_i - \mathbf{p}||_2^2}
$$
 where $\gamma$ is the coverage radius of a node, for which we set to 0.9 cm for 4DMatch examples, $C$ denotes the normalization constant, ensuring that skinning weights add up to one 
$$
\sum_{\mathcal{V}_i \in \mathcal{V}} w_{\mathbf{p},i} = 1
$$

\medskip
\noindent
\textbf{Energy Function}. The energy function of non-rigid iterative closest point (N-ICP) consists of two terms: the correspondence term and the regularization term.
Given a set of matches $\mathcal{K}$, and the confidence of the correspondences  $ c_{(\mathbf{p}_s,\mathbf{p}_t)} $ where $ (\mathbf{p}_s,\mathbf{p}_t) \in \mathcal{K} $.
Correspondence term is defined as 
$$
\mathbf{E}_{corr} (\mathcal{G}) =\sum_{(\mathbf{p}_s,\mathbf{p}_t) \in \mathcal{K}} 
c _{(\mathbf{p}_s,\mathbf{p}_t)}^2
|| \mathcal{W}(\mathbf{p}_s) - \mathbf{p}_t ||_2^2 
$$
We use ARAP~\cite{arap} as the regularization term 
$$
\mathbf{E}_{reg}(\mathcal{G}) =\sum_{(i,j)\in \mathcal{E}}   || 
      \mathbf{R}_i(\mathbf{g}_j -\mathbf{g}_i ) + 
      \mathbf{g}_i + \mathbf{t}_i -   (\mathbf{g}_j + \mathbf{t}_j)  
      ||^2
$$
The total energy function is
$$
\mathbf{E}_{reg}(\mathcal{G}) =\lambda_c \mathbf{E}_{corr}(\mathcal{G}) +\lambda_a\mathbf{E}_{reg}(\mathcal{G})
$$

\medskip
\noindent
\textbf{Residual and Partial Derivatives.} The followings show the residuals and partial derivatives for optimization. Derivative of the wrapping function $\mathcal{W}$
$$
\frac{\partial \mathcal{W}(\mathbf{p})}{\partial \mathbf{\varphi}_i} 
=  -w_{\mathbf{p},i} (\mathbf{R}_i (\mathbf{p}-\mathbf{g}_i))^{\wedge}
$$$$
\frac{\partial \mathcal{W}(\mathbf{p})}{\partial \mathbf{t}_i} 
=  w_{\mathbf{p},i} \mathbf{I}_3
$$where $\mathbf{I}_3$ is the $3\times3$ identity matrix. 
Residual term for a correspondence $(\mathbf{p}_s,\mathbf{p}_t) \in \mathcal{K}$
$$
\mathbf{r}^{corr}_{(\mathbf{p}_s,\mathbf{p}_t)}
= 
\sqrt{ \lambda_c} c _{(\mathbf{p}_s,\mathbf{p}_t)}
( \mathcal{W}(\mathbf{p}_s) - \mathbf{p}_t ) 
$$Derivative of correspondence residual $\mathbf{r}^{corr}_{(\mathbf{p}_s,\mathbf{p}_t)}$ $$
\frac{ \partial \mathbf{r}^{corr}_{(\mathbf{p}_s,\mathbf{p}_t)}}{\partial \mathbf{\varphi}_i} 
=  - \sqrt{ \lambda_c} c _{(\mathbf{p}_s,\mathbf{p}_t)} w_{\mathbf{p}_s,i}  (\mathbf{R}_i (\mathbf{p}_s-\mathbf{g}_i))^{\wedge}
$$$$\frac{ \partial \mathbf{r}^{corr}_{(\mathbf{p}_s,\mathbf{p}_t)}}{\partial \mathbf{t}_i} 
=  \sqrt{ \lambda_c} c _{(\mathbf{p}_s,\mathbf{p}_t)} w_{\mathbf{p},i} \mathbf{I}_3
$$Residual term for regularization tern $(i,j) \in \mathcal{E}$
$$
\mathbf{r}^{reg}_{(i,j)}
=
\sqrt{ \lambda_c} ( \mathbf{R}_i(\mathbf{g}_j -\mathbf{g}_i ) +  \mathbf{g}_i + \mathbf{t}_i -   (\mathbf{g}_j + \mathbf{t}_j) )
$$Derivative of regularization term $\mathbf{r}^{reg}_{(i,j)}$
$$
\frac{ \partial \mathbf{r}^{reg}_{(i,j)}}{\partial \mathbf{\varphi}_i} 
=  - \sqrt{ \lambda_a} (\mathbf{R}_i (\mathbf{g}_j-\mathbf{g}_i))^{\wedge}
$$$$
\frac{ \partial \mathbf{r}^{reg}_{(i,j)}}{\partial \mathbf{t}_i} 
=  \sqrt{ \lambda_a} \mathbf{I}_3
$$$$
\frac{ \partial \mathbf{r}^{reg}_{(i,j)}}{\partial \mathbf{t}_j} 
=  -\sqrt{ \lambda_a}\mathbf{I}_3
$$
The full Jacobian matrix $\mathbf{J} \in \mathbb{R}^{ (3|\mathcal{K}|+3|\mathcal{E}|) \times 6|\mathcal{V}|}$ is shown as 
\begin{center}
$
\mathbf{J}=
$
\scalebox{0.9}{
$
\renewcommand{\arraystretch}{2.5}
\begin{array}{|ccc|ccc|}

\hline

 \frac{ \partial \mathbf{r}^{corr}_{1}}{\partial \mathbf{\varphi}_1}
&  \cdots 
& \frac{ \partial \mathbf{r}^{corr}_{1}}{\partial \mathbf{\varphi}_{|\mathcal{V}|}} 
& \frac{ \partial \mathbf{r}^{corr}_{1}}{\partial \mathbf{t}_1}  
&  \cdots 
& \frac{ \partial \mathbf{r}^{corr}_{1}}{\partial \mathbf{t}_{|\mathcal{V}|}}  \\   

 \vdots & \ddots  & \vdots &  \vdots & \ddots  & \vdots \\ 

 \frac{ \partial \mathbf{r}^{corr}_{|\mathcal{K}|}}{\partial \mathbf{\varphi}_1} 
&  \cdots
& \frac{ \partial \mathbf{r}^{corr}_{|\mathcal{K}|}}{\partial\mathbf{\varphi}_{|\mathcal{V}|}}
& \frac{ \partial \mathbf{r}^{corr}_{|\mathcal{K}|}}{\partial \mathbf{t}_1}  
 &  \cdots
& \frac{ \partial \mathbf{r}^{corr}_{|\mathcal{K}|}}{\partial \mathbf{t}_{|\mathcal{V}|}}  \\   \hline

 \frac{ \partial \mathbf{r}^{reg}_{1}}{\partial \mathbf{\varphi}_1} 
&  \cdots 
& \frac{ \partial \mathbf{r}^{reg}_{1}}{\partial \mathbf{\varphi}_{|\mathcal{V}|}} 
& \frac{ \partial \mathbf{r}^{reg}_{1}}{\partial \mathbf{t}_1}  
&  \cdots 
& \frac{ \partial \mathbf{r}^{reg}_{1}}{\partial \mathbf{t}_{|\mathcal{V}|}}  \\

 \vdots & \ddots  & \vdots &  \vdots & \ddots  & \vdots \\ 

 \frac{ \partial \mathbf{r}^{reg}_{|\mathcal{E}|}}{\partial \mathbf{\varphi}_1} 
&  \cdots
& \frac{ \partial \mathbf{r}^{reg}_{|\mathcal{E}|}}{\partial \mathbf{\varphi}_{|\mathcal{V}|}} 
& \frac{ \partial \mathbf{r}^{reg}_{|\mathcal{E}|}}{\partial \mathbf{t}_1}  
&  \cdots
& \frac{ \partial \mathbf{r}^{reg}_{|\mathcal{E}|}}{\partial \mathbf{t}_{|\mathcal{V}|}}

\\ \hline 
\end{array}
$
}
\end{center}

where $|\mathcal{V}|$ is the number of graph node.  $|\mathcal{K}|$ is number of correspondence. $|\mathcal{E}|$ is the number of graph edge. Each block in $\mathbf{J}$ is a $3\times3$ matrix. For the sparse nature of this problem, most blocks are zeros.
The full residual vector $\mathbf{r}\in \mathbb{R}^{3|\mathcal{K}|+3|\mathcal{E}|}$ is shown as

\begin{center}
\scalebox{0.9}{$
\mathbf{r} =
\renewcommand{\arraystretch}{2.5}
\begin{array}{|c|}
\hline
\mathbf{r}_{1}^{corr} \\
\vdots \\
\mathbf{r}_{|\mathcal{K}|}^{corr}  \\
\mathbf{r}_{1}^{reg} \\
\vdots \\
\mathbf{r}_{|\mathcal{E}|}^{reg} \\
\hline
\end{array}
$
}
\end{center}

where each block is a $3\times1$ vector. The total length is $(|\mathcal{K}| + |\mathcal{E}|)\times3$.

\medskip
\noindent
\textbf{Non-rigid Optimization.}
We use Gauss-Newton algorithm and minimizes the total energy function $\mathbf{E}_{total}$.
The Gauss-Newton method is an iterative scheme. In every iteration $n$, we re-compute the Jacobian matrix $\mathbf{J}$ and the residual vector $\mathbf{r}$ , and get a solution increment $\Delta \mathcal{G}$ by solving the update equations:
$$
\mathbf{J}^T\mathbf{J}\Delta\mathcal{G} = \mathbf{J}^T\mathbf{r}
$$
The above linear system is solved using LU decomposition.

\section{Qualitative Results}\label{sec:s-qualitative}

\begin{figure*}
    \centering
    \includegraphics[width=0.85\linewidth]{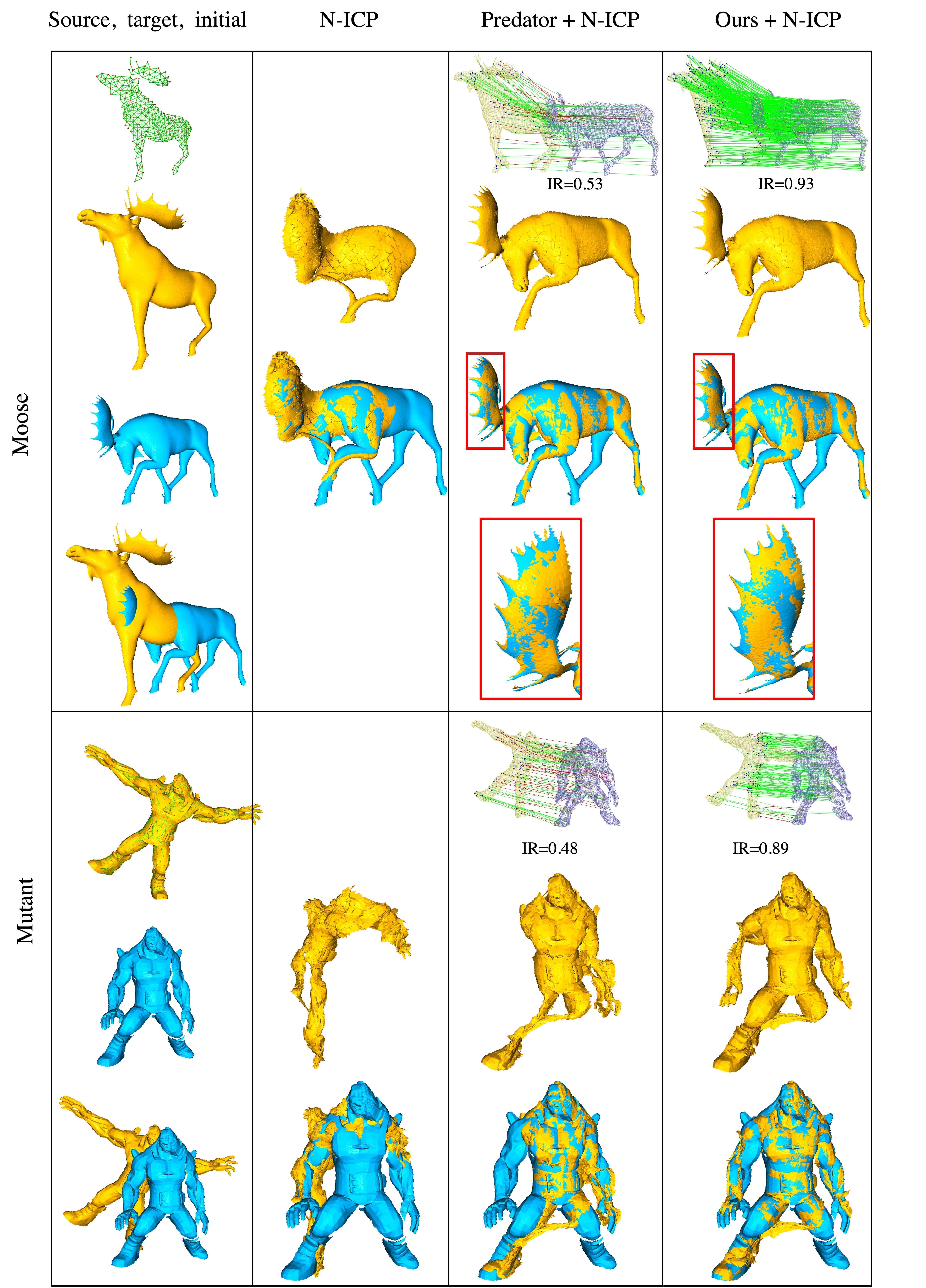}
    \caption{Qualitative point cloud matching and registration results on 4DMatch.
    The inlier threshold is set to $4cm$. 
    The N-ICP-based refinement can remedy outliers to a certain extent if the outlier matches are not too far away from the ground truth (see the results of \textit{Predator + N-ICP} in the Moose example). 
    The N-ICP-based refinement can not handle outliers that connects distant parts. E.g. in the Mutant example, left and right legs are registered together by both methods.
    }
    \label{fig:4dreg}
\end{figure*}

\begin{figure*}
    \centering
    \includegraphics[width=1\linewidth]{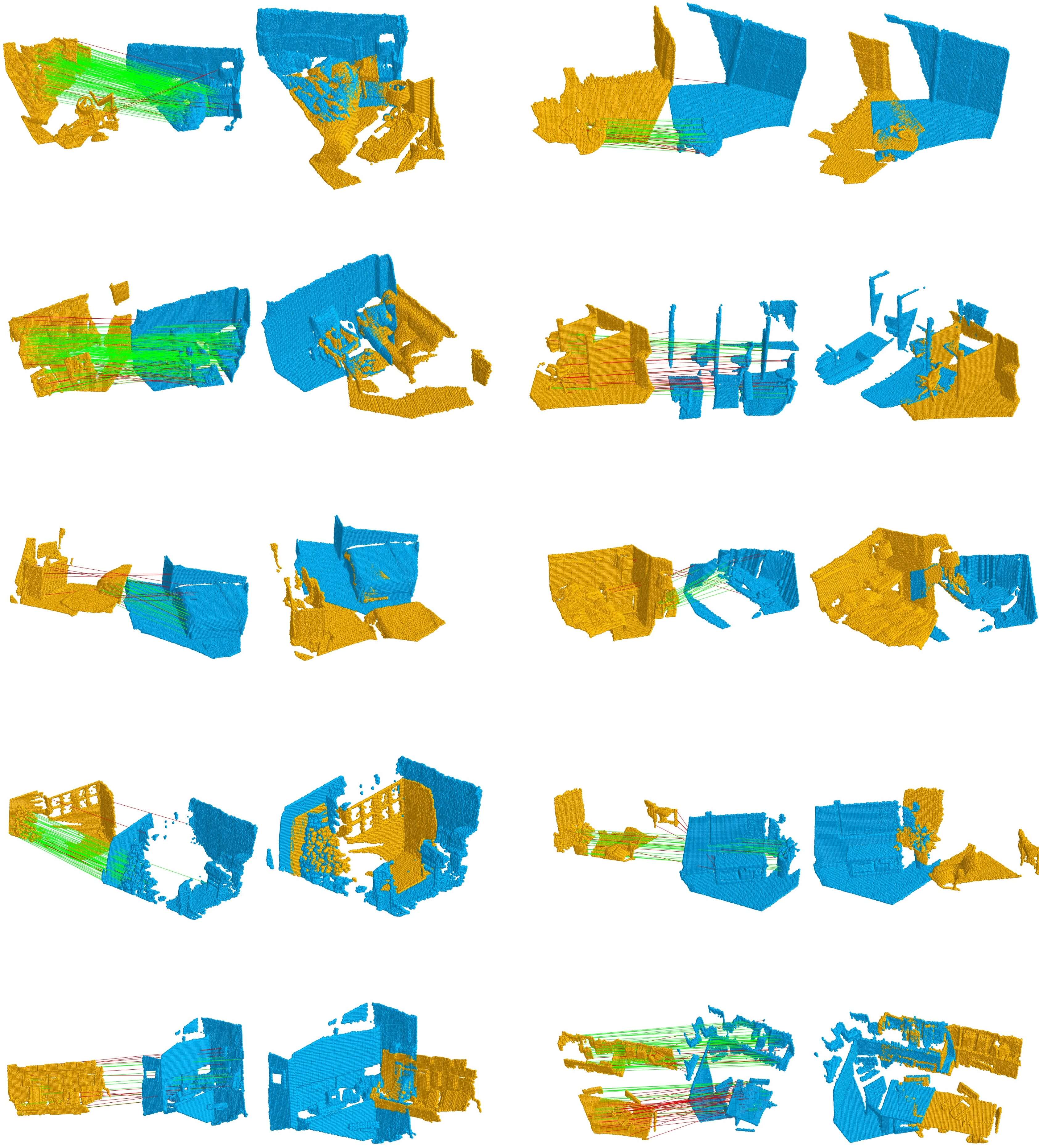}
    \caption{Qualitative point cloud matching and registration results on 3DLoMatch.}
    \label{fig:3dreg}
\end{figure*}

Tab.~\ref{tab:end_point_error} shows the scores for the elephant and dragon examples from the main paper.
Fig.~\ref{fig:4dreg} shows the qualitative matching and registration results on 4DMatch. 
Tab.~\ref{tab:s-end_point_error} shows the corresponding scores for results in Fig.~\ref{fig:4dreg}.
Fig.~\ref{fig:3dreg} shows the qualitative matching and registration results on 3DLoMatch.

\begin{table}[!ht]

\newcolumntype{L}{>{\centering\arraybackslash}m{2cm}}
\centering
\resizebox{1\columnwidth}{!}{
\begin{tabular}{l|ccc|ccc}
\toprule

&\multicolumn{3}{c|}{elephant}&\multicolumn{3}{c}{dragon}\\
&EPE$\downarrow$&Acc5$\uparrow$&Acc10$\uparrow$&EPE$\downarrow$&Acc5$\uparrow$&Acc10$\uparrow$\\
\midrule
N-ICP&0.166&22.4&41.5&0.325&4.5&17.8\\
Predator\cite{huang2021predator} + N-ICP&0.092&55.7&66.0&0.514&29.6&32.1\\
Ours + N-ICP&\textbf{0.018}&\textbf{90.6}&\textbf{98.0}&\textbf{0.038}&\textbf{68.0}&\textbf{96.0}\\
\bottomrule
\end{tabular}
}
\caption{Quantitative non-rigid registration results. 
The metrics are 3D end point error (EPE) and motion estimation accuracy (Acc) ($<$0.05$m$ or
5\%, $<$0.1$m$ or 10\%).
}
\label{tab:end_point_error}
\end{table}

\begin{table}[!h]

\newcolumntype{L}{>{\centering\arraybackslash}m{2cm}}
\centering
\resizebox{1\columnwidth}{!}{
\begin{tabular}{l|ccc|ccc}
\toprule

&\multicolumn{3}{c|}{moose}&\multicolumn{3}{c}{mutant}\\
&EPE$\downarrow$&Acc5$\uparrow$&Acc10$\uparrow$&EPE$\downarrow$&Acc5$\uparrow$&Acc10$\uparrow$\\
\midrule
N-ICP&0.728&0.1&0.7&0.52&0.0&0.6\\
Predator\cite{huang2021predator} + N-ICP&0.0283& 86.9&99.4&0.217&44.6&60.1\\
Ours + N-ICP&\textbf{0.0263}&\textbf{88.5}&\textbf{99.9}&\textbf{0.119}&\textbf{62.6}&\textbf{71.4}\\
\bottomrule
\end{tabular}
}
\caption{Quantitative non-rigid registration results. 
The metrics are 3D end point error (EPE) and motion estimation accuracy (Acc) ($<$0.05$m$ or
5\%, $<$0.1$m$ or 10\%).
}
\label{tab:s-end_point_error}
\end{table}

\end{document}